\documentclass[journal]{IEEEtran}
\ifCLASSINFOpdf
\else
\fi
\usepackage{times}
\usepackage{epsfig}
\usepackage{graphicx}
\usepackage{amsmath}
\usepackage{dsfont}
\usepackage{amssymb}
\usepackage{url}
\usepackage{amsfonts}
\usepackage{subfigure}
\usepackage{multirow}
\usepackage{makecell}
\usepackage{booktabs}
\usepackage{lineno}
\usepackage{graphicx}
\usepackage{caption}
\begin{document}
\title{A Comprehensive Survey of 3D Dense Captioning: Localizing and Describing Objects in 3D Scenes}
\author{Ting~Yu,~\IEEEmembership{Member,~IEEE},
       Xiaojun~Lin,
       Shuhui~Wang,~\IEEEmembership{Member,~IEEE},
       Weiguo~Sheng,~\IEEEmembership{Member,~IEEE},
       Qingming~Huang,~\IEEEmembership{Fellow,~IEEE}, and
       Jun~Yu,~\IEEEmembership{Senior Member,~IEEE}
}

\twocolumn[{
\renewcommand\twocolumn[1][]{#1}
\maketitle

\begin{center}
   \captionsetup{type=figure}
   \includegraphics[width=1\textwidth]{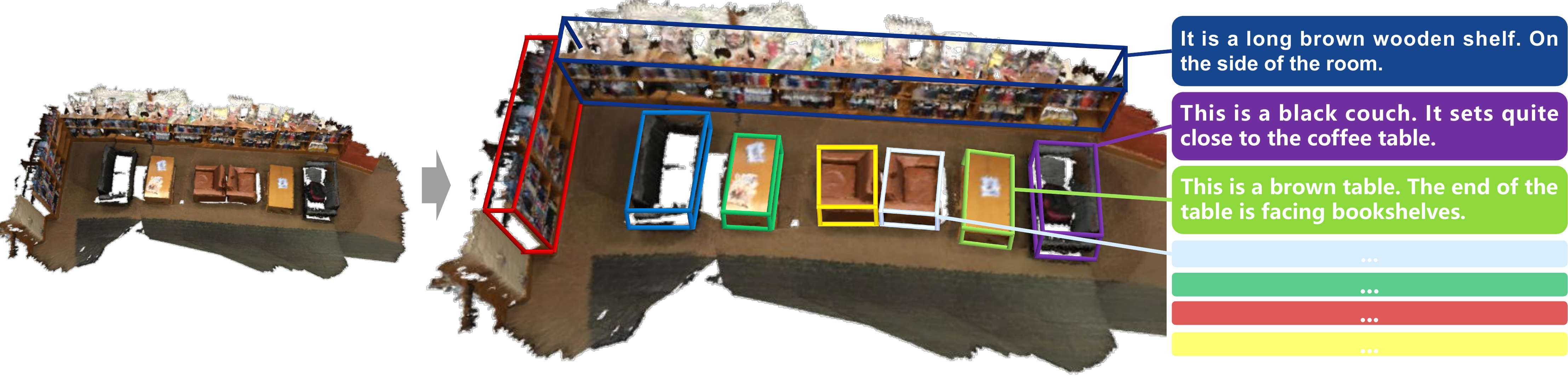}
   \captionof{figure}{Illustration of a 3D dense captioning task: localizing and describing objects in 3D scenes. The task involves the combined process of object localization and captioning to generate natural language descriptions for objects in a 3D scene. It takes a 3D point cloud source as input and produces a diverse range of bounding boxes along with multiple descriptions.}
\label{fig1}
\end{center}}]

\let\thefootnote\relax\footnotetext{This work was supported by the National Natural Science Foundation of China under Grant No. 62002314, 62125201, 62020106007, 62022083, 62236008 and Zhejiang Provincial Natural Science Foundation of China under Grant No. LY23F020005. (Corresponding author: Jun Yu.)}
\let\thefootnote\relax\footnotetext{T. Yu, X. Lin, and W. Sheng are with the School of Information Science and Technology, Hangzhou Normal University, Hangzhou 311121, China (e-mail: yut@hznu.edu.cn; linxiaojun@stu.hznu.edu.cn; w.sheng@ieee.org).}
\let\thefootnote\relax\footnotetext{S. Wang is with the Key Laboratory of Intelligent Information Processing, Institute of Computing Technology, Chinese Academy of Sciences, Beijing 100190, China (e-mail: wangshuhui@ict.ac.cn).}
\let\thefootnote\relax\footnotetext{Q. Huang is with the School of Computer Science and Technology, University of Chinese Academy of Sciences, Beijing 101408, China (e-mail: qmhuang@ucas.ac.cn).}
\let\thefootnote\relax\footnotetext{J. Yu is with the Key Laboratory of Complex Systems Modeling and Simulation, School of Computer Science and Technology, Hangzhou Dianzi University, Hangzhou 310018, China (email: yujun@hdu.edu.cn).}

\begin{abstract}
Three-Dimensional (3D) dense captioning is an emerging vision-language bridging task that aims to generate multiple detailed and accurate descriptions for 3D scenes. It presents significant potential and challenges due to its closer representation of the real world compared to 2D visual captioning, as well as complexities in data collection and processing of 3D point cloud sources. Despite the popularity and success of existing methods, there is a lack of comprehensive surveys summarizing the advancements in this field, which hinders its progress.
In this paper, we provide a comprehensive review of 3D dense captioning, covering task definition, architecture classification, dataset analysis, evaluation metrics, and in-depth prosperity discussions. Based on a synthesis of previous literature, we refine a standard pipeline that serves as a common paradigm for existing methods. We also introduce a clear taxonomy of existing models, summarize technologies involved in different modules, and conduct detailed experiment analysis. Instead of a chronological order introduction, we categorize the methods into different classes to facilitate exploration and analysis of the differences and connections among existing techniques. We also provide a reading guideline to assist readers with different backgrounds and purposes in reading efficiently.
Furthermore, we propose a series of promising future directions for 3D dense captioning by identifying challenges and aligning them with the development of related tasks, offering valuable insights and inspiring future research in this field. Our aim is to provide a comprehensive understanding of 3D dense captioning, foster further investigations, and contribute to the development of novel applications in multimedia and related domains.
\end{abstract}

\begin{IEEEkeywords}
3D dense captioning, vision-language bridging, visual captioning, 3D point cloud.
\end{IEEEkeywords}

\section{Introduction}
\IEEEPARstart{H}umans possess a remarkable ability to rapidly recognize and describe various shape details and spatial relationships of objects in unfamiliar scenarios with just a brief glimpse \cite{fei2007we}. However, replicating this capability in current computer systems remains challenging. Previous influential research in related fields has predominantly focused on the task of image captioning \cite{9724218,9784827,8869845,vinyals2015show,xu2015show,lu2017knowing,anderson2018bottom}, which involves bridging visual content understanding with natural language description \cite{vaswani2017attention,devlin2018bert} to generate captions that highlight the overall content of the entire image. Subsequently, dense image captioning \cite{karpathy2015deep,johnson2016densecap,yang2017dense,yin2019context,kim2019dense} emerged as a natural extension of image captioning, witnessing a surge of interest in cross-media unified expression facilitated by advances in deep learning technology. Unlike image captioning, dense image captioning places greater emphasis on identifying and expressing local visual details in multiple natural languages.

Despite the widespread popularity and significant success of the aforementioned captioning tasks, they are not without limitations. One of the main limitations is that they rely solely on 2D image sources, which inherently have a single viewpoint and can result in misaligned, distorted, and obscured appearances, making it challenging to capture comprehensive physical details and object location relationships \cite{ghandi2022deep}. However, more recently, with the advent of 3D point cloud data collection techniques \cite{armeni20163d,hua2016scenenn,dai2017scannet} and pioneering point cloud processing methods \cite{qi2017pointnet++,qi2019deep,hou2020revealnet}, there has been a growing interest in 3D cross-modal learning. Unlike 2D images with uniform pixels and grids, 3D point clouds are represented by sparse and disordered points, as illustrated in Fig. \ref{fig3}, and provide rich geometric information, including physical characteristics such as size and shape, as well as spatial relationships between objects from multiple rotatable perspectives. This property of 3D point clouds significantly compensates for the limitations of 2D visual data. Consequently, Chen \emph{et al.} \cite{chen2021scan2cap} proposed 3D dense captioning as a novel task for generating dense captions in 3D scenes, elevating the traditional dense captioning task from 2D to 3D and resulting in more customized and detailed descriptions. As a burgeoning vision-language-bridging task, 3D dense captioning targets to jointly localize and describe individual objects in 3D scenes leveraging commodity RGB-D sensors. The input for this task is the point cloud of 3D scenes, and the output includes descriptions for the underlying objects along with specific bounding boxes, as shown in Fig. \ref{fig1}. Furthermore, 3D dense captioning can be further divided into two sub-tasks: object detection and object caption generation.

Due to the success of the pioneering work \cite{chen2021scan2cap}, the research on 3D dense captioning has been rolled out comprehensively and rapidly in the following years. Several notable approaches have been proposed to address key challenges in 3D dense captioning \cite{jiao2022more,wang2022spatiality,yuan2022x,cai20223djcg,takahashi2020d3net,zhong2022contextual,chen2023end}. For instance, MORE \cite{jiao2022more}, SpaCap3D \cite{wang2022spatiality}, and Scan2Cap \cite{chen2021scan2cap} employed different relation reasoning modules to enhance the relationships among candidate objects. X-Trans2Cap \cite{yuan2022x} introduced a multi-modal knowledge transfer network with 2D priors to improve 3D dense captioning. 3DJCG \cite{cai20223djcg} and D3Net \cite{takahashi2020d3net} focused on the connection between 3D dense captioning and 3D visual grounding \cite{chen2020scanrefer,yuan2021instancerefer,zhao20213dvg}. CM3D \cite{zhong2022contextual} aimed to incorporate contextual knowledge from point clouds into 3D dense captioning. In contrast, Vote2Cap-DETR \cite{chen2023end} proposed a parallel approach that combines localization and captioning, deviating from the traditional \emph{detect-then-describe} pipeline. The development timeline for 3D dense captioning is illustrated in Fig. \ref{fig7}. 

Recent advancements highlight the active and ongoing research efforts, striving to overcome challenges and push the boundaries of 3D dense captioning.
However, the existing literature lacks a comprehensive survey on the topic of 3D dense captioning, despite its increasing prevalence. Therefore,  this paper aims to bridge this gap by providing a comprehensive and insightful overview that bridges past research with future prospects in the field of 3D dense captioning. The overview will encompass critical components, including task introduction, methodology review, and future outlook, with the purpose of offering valuable insights for researchers and practitioners. The major contributions can be summarized as the following three aspects:
\begin{figure}[h]
  \centering
   \subfigure[2D Image] {\includegraphics[width=0.47\linewidth]{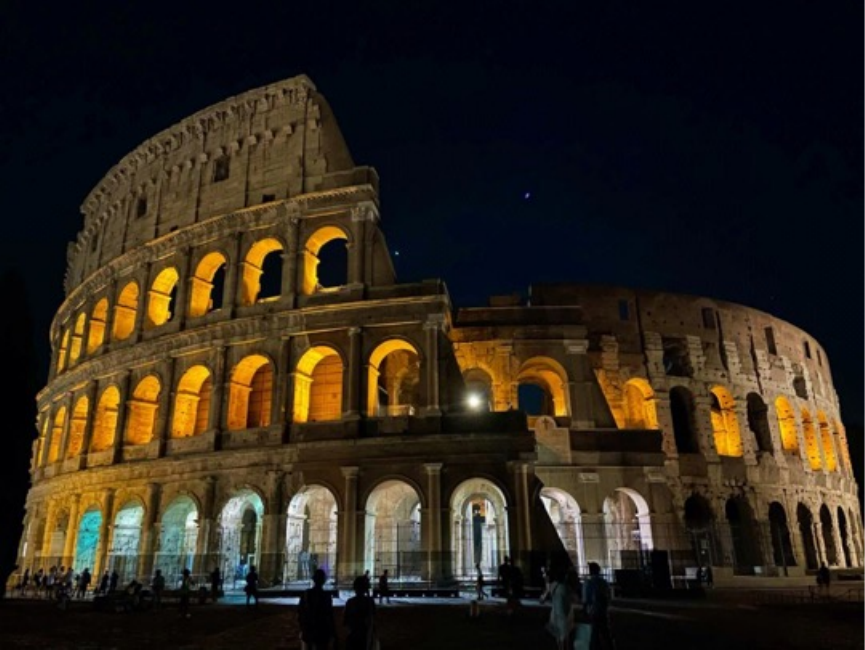}\label{fig:2d}}
   \subfigure[3D Point Cloud] {\includegraphics[width=0.51\linewidth]{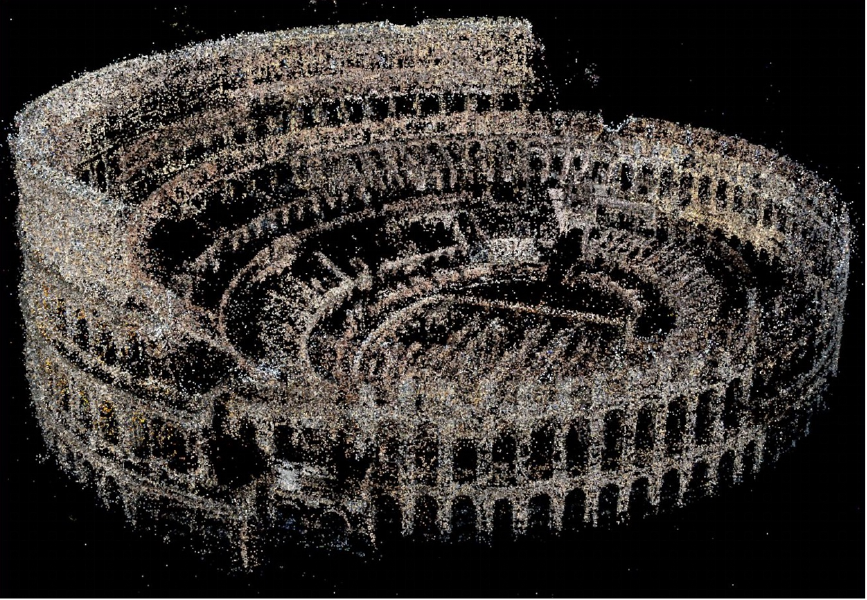}\label{fig:3d}}
  \caption{Visualization of 3D point cloud and 2D image. The inherent single viewpoint of 2D images inevitably triggers a misaligned, distorted, and obscured appearance. Compared to 2D images with uniform pixels and grids, 3D point clouds are represented by sparse and disordered points, providing comprehensive geometric information, including physical characteristics such as size and shape and prosperous spatial relationships from multiple rotatable perspectives. }
 \label{fig3}
\end{figure}

\begin{itemize}
\item Comprehensive and insightful review: This paper presents the first known survey that offers a comprehensive and insightful review of 3D dense captioning. It covers various aspects, such as task definition, architecture classification, datasets analysis, evaluation metrics, and multi-faceted discussions, providing a holistic understanding of 3D dense captioning.
\item Critical analysis of existing architectures: Instead of a chronological order introduction, this paper categorizes the existing methods into different classes, enabling a more beneficial exploration and analysis of the differences and connections among the models. This critical analysis provides valuable insights into the strengths and limitations of current approaches, aiding researchers and practitioners in making informed decisions.
\item Proposal of future directions: Drawing upon the challenges identified in the field, this paper proposes a series of promising future directions for 3D dense captioning. These directions are aligned with the developments in related tasks, aiming to inspire future research endeavors and drive further advancements in the field.
\end{itemize}

\begin{figure*}[h]
  \centering
  \includegraphics[width=\linewidth]{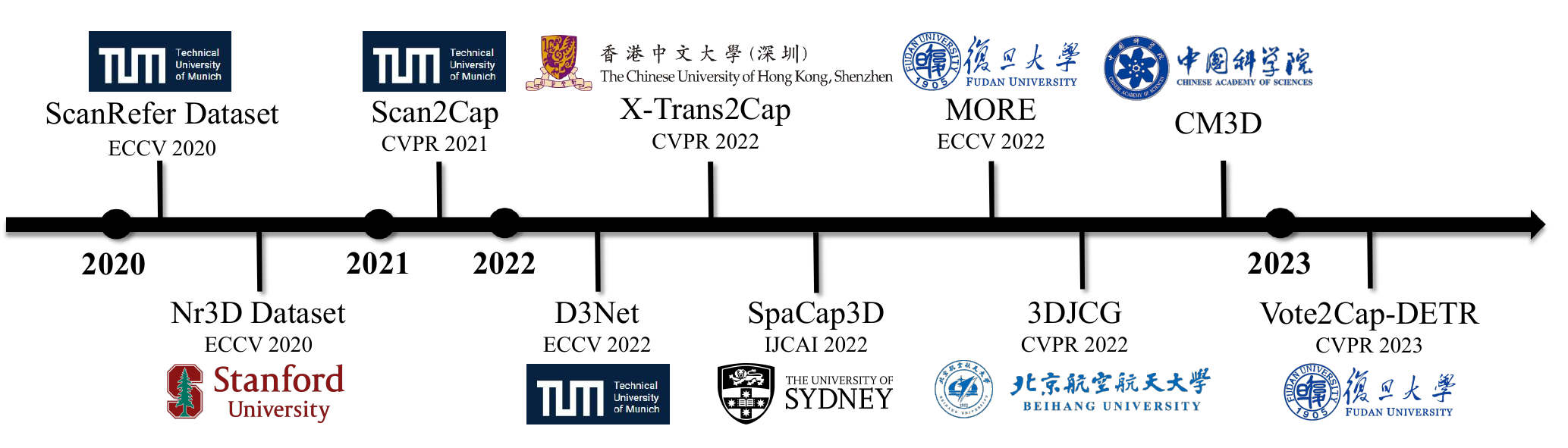}
 \caption{Evolution of 3D dense captioning: datasets and models over time. Two fundamental datasets ScanRefer \cite{chen2020scanrefer} and Nr3D \cite{achlioptas2020referit3d}, have played a pivotal role in shaping the field. The ScanRefer dataset was initially curated for the 3D visual grounding task, and Nr3D is a sub-dataset of ReferIt3D \cite{achlioptas2020referit3d}, comprising human-annotated 3D scenes. In subsequent years, the field witnessed a surge of novel models, including Scan2Cap \cite{chen2021scan2cap}, D3Net \cite{takahashi2020d3net}, X-Trans2Cap \cite{yuan2022x}, SpaCap3D \cite{wang2022spatiality}, MORE \cite{jiao2022more}, 3DJCG \cite{cai20223djcg}, CM3D \cite{zhong2022contextual}, and Vote2Cap-DETR \cite{chen2023end}. Notably, the order of these models is based on the commit dates on the benchmark, as the exact appearance times of papers were ambiguous.}
\label{fig7}
\end{figure*}

The paper is organized into four main sections to ensure a clear and organized presentation. In Section \ref{sec:reading guideline}, a targeted reading guideline is provided to assist readers with different backgrounds and purposes. Section \ref{sec:related_work} discusses the four related tasks, including image captioning, dense image captioning, dense video captioning, and 3D visual grounding. In Section \ref{sec:architecture}, we synthesize task definition, main framework, and model classification, the most substantial and crucial components of the paper. Section \ref{sec:task} introduces 3D dense captioning regarding dataset analysis and evaluation metrics. In Section \ref{sec:exam}, we analyze the experimental details, including the loss function and model performance. Furthermore, the challenges of past 3D dense captioning techniques are discussed, and potential future innovations are proposed in Section \ref{sec:disscussion}. Finally, we conclude this survey in Section \ref{sec:conclusion}.

\section{reading guidelines}\label{sec:reading guideline}
This paper aims to provide comprehensive insights into the field of 3D dense captioning, catering to readers with varying backgrounds and interests. To optimize the utilization of this paper, the following reading guidance is provided:

\begin{itemize}
\item Beginners in the 3D visual-language domain: If readers are new to the field and lack prior experience in related areas, we recommend reading the entire paper section by section. This approach will facilitate a comprehensive understanding of the landscape of 3D dense captioning, as the paper covers relevant literature in detail.
\item Researchers familiar with related tasks: Readers with prior experience in tasks such as image captioning or 3D grounding may selectively skip some sections. For instance, the dataset analysis in Section \ref{sec:task} may be familiar to those working on 3D grounding tasks.
\item Experienced researchers of 3D dense captioning: If readers are already well-versed in the field of 3D dense captioning, Sections IV and VII may be particularly meaningful to them. These sections summarize existing models and provide insights into future directions, which may be of interest to researchers with expertise in the field.
\item Readers with specific goals: Readers with clear goals or motivations for reading this paper can directly jump to the corresponding section they want to focus on, as each section is relatively independent. This approach allows for quick access to the specific information needed without reading the entire paper.
\end{itemize}

We hope that this reading guideline will assist readers in maximizing the benefits of this paper and enhancing their understanding of 3D dense captioning, a multimodal learning task in the domain of 3D scene understanding.

\begin{figure*}[h]
  \centering
  \includegraphics[width=\linewidth]{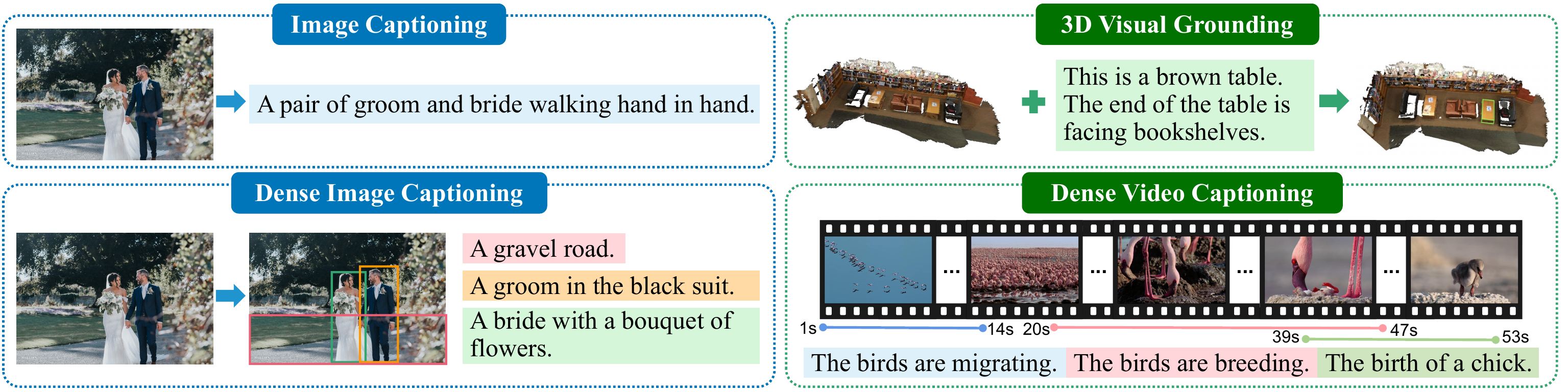}
   \caption{Illustration of representative visual-language bridging tasks related to 3D dense captioning. Both image captioning and dense image captioning involve using a 2D image as input to generate natural language descriptions. However, they differ in their objectives. Image captioning aims to generate a sentence that describes the overall content of the input image. On the other hand, dense captioning, as a variation of image captioning, focuses on generating distinctive descriptions for prominent regions within an image, capturing fine-grained details. 3D visual grounding is a task that emphasizes localizing the object described by an input sentence within a 3D scene. It involves linking the language-based description to the corresponding 3D object in the scene, bridging the gap between visual perception and natural language understanding. Dense video captioning involves localizing significant events in an untrimmed video and generating captions for each event. Among the four tasks mentioned, dense image captioning exhibits the closest resemblance to 3D dense captioning. Conversely, 3D visual grounding stands in contrast to 3D dense captioning, emphasizing different aspects of visual understanding and language integration.}
\label{fig2}
\end{figure*}

\section{Related Work}\label{sec:related_work}
In this section, we briefly review the most relevant research on 3D dense captioning, especially those representative studies that involve captioning or 3D fields, including image captioning, dense image captioning, dense video captioning, and 3D visual grounding, as illustrated in Fig. \ref{fig2}.

\subsection{Image Captioning}
As a representative visual-language generation task \cite{9381876,9521159}, image captioning aims to provide a descriptive sentence for an input image, with the purpose of facilitating 2D scene comprehension, particularly for individuals with visual impairments \cite{hossain2019comprehensive}. The development of image captioning can be delineated into two distinct phases: the traditional machine learning stage and the advent of deep learning approaches, as discussed in \cite{ghandi2022deep, senior2023graph}. In the earlier machine learning stage, template-based \cite{farhadi2010every} and retrieval-based \cite{wu2017automatic} methods were commonly employed. However, with the emergence of deep learning \cite{ghandi2022deep}, researchers have shifted their focus towards leveraging advanced technologies, such as attention mechanisms \cite{anderson2018bottom, xu2015show, yu2019compositional}, graph neural networks \cite{gu2019scene, xu2017scene, yang2019auto}, convolutional networks \cite{girshick2015fast, ren2015faster, kipf2016semi}, transformers \cite{vaswani2017attention, herdade2019image, cornia2020meshed}, and Vision-Language Pre-training (VLP) models \cite{radford2021learning, radford2019language}.
Typically, the core encoder-decoder framework has been popularly used in existing image captioning approaches. Convolutional Neural Networks (CNNs) \cite{krizhevsky2017imagenet, he2016deep, huang2017densely} were employed as encoders to map input images into feature vectors, while Recurrent Neural Networks (RNNs) \cite{hochreiter1997long, cho2014learning} were utilized as decoders to generate sentences from the feature vectors. Furthermore, some approaches \cite{cornia2020meshed, huang2019attention, huang2019adaptively, wang2020show, yu2020long} incorporated object detectors to enhance the extraction of visual features.
Attention mechanisms \cite{cornia2020meshed, pan2020x, schwartz2017high} and graph neural networks \cite{yang2019auto, yao2018exploring} have gained increasing popularity in recent years due to their ability to capture fine-grained visual details and contextual relationships. Notably, methods incorporating VLP models, such as ConZIC \cite{zeng2023conzic}, have demonstrated efficient performance in image captioning. As a self-correction framework, ConZIC integrated BERT \cite{devlin2018bert} and CLIP \cite{radford2021learning}, enabling controllable zero-shot image captioning with significantly improved generation speed compared to previous works.

\subsection{Dense Image Captioning}
As a specialized type of image captioning, dense image captioning focuses on generating separate descriptions for each prominent region or object within an image, closely related to 3D dense captioning \cite{chen2021scan2cap}. The advent of \cite{johnson2016densecap} in 2016 marked a significant milestone in the field of dense image captioning. In contrast to global-based image captioning, which provides a single caption for the entire image, dense image captioning generates more detailed and region-level descriptions for multiple objects or regions in the image. To address the challenge of incorporating non-object context in prior work, Yang \emph{et al.} proposed a context fusion module to generate more accurate and contextually relevant descriptions \cite{yang2017dense}. Context fusion module helps to improve the quality of the generated descriptions by considering the surrounding visual elements beyond individual objects.
Furthermore, Song \emph{et al.} presented a semantically symmetric LSTM \cite{hochreiter1997long} model combining dense image captioning with scene recognition models \cite{song2019much}. It leveraged the relationship between scene understanding and image captioning to generate more meaningful and coherent captions that capture both the local object-level details and the global scene context. More recently, there has been a surge of interest in multi-modal pre-trained methods for joint dense image captioning and object detection, leading to a new wave of research in this area. Prominent models \cite{long2023capdet,gao2022caponimage} integrated object detection and dense captioning into a unified framework for improved performance and computation efficiency. These approaches leveraged pre-trained models that incorporate multiple modalities, such as images and text, to jointly learn the representations for both tasks, leading to promising advancements in the field of dense image captioning.

\begin{figure*}[h]
  \centering
  \includegraphics[width=\linewidth]{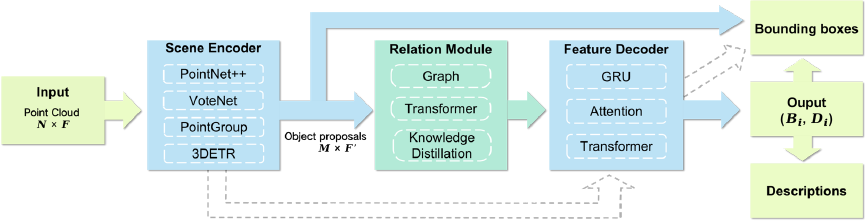}
\caption{The flowchart of the general framework for 3D dense captioning, which typically encompasses three main components: a scene encoder, a relation module, and a feature decoder. The relation module is a core component that is commonly employed in most existing works, rendering it an integral part of the encoder-decoder structure. Specifically, the input to the framework is a 3D point cloud containing $N$ randomly sampled points, each characterized by $F$-dimensional features. The model then generates $I$ pairs of bounding boxes along with multiple descriptions $(B_i, D_i)$ as outputs. In the intermediate process, most models generate $M$ object proposals with $F'$-dimensional feature representations using a scene encoder. These proposals are then utilized to learn the relationships between objects in the relation module. Finally, the feature decoder generates corresponding descriptions for the objects. Notably, a smaller portion of methods do not follow this approach and instead perform detection and captioning simultaneously in the final feature encoding step.}
\label{flowchart}
\end{figure*}

\begin{figure}
  \centering
  \includegraphics[width=\linewidth]{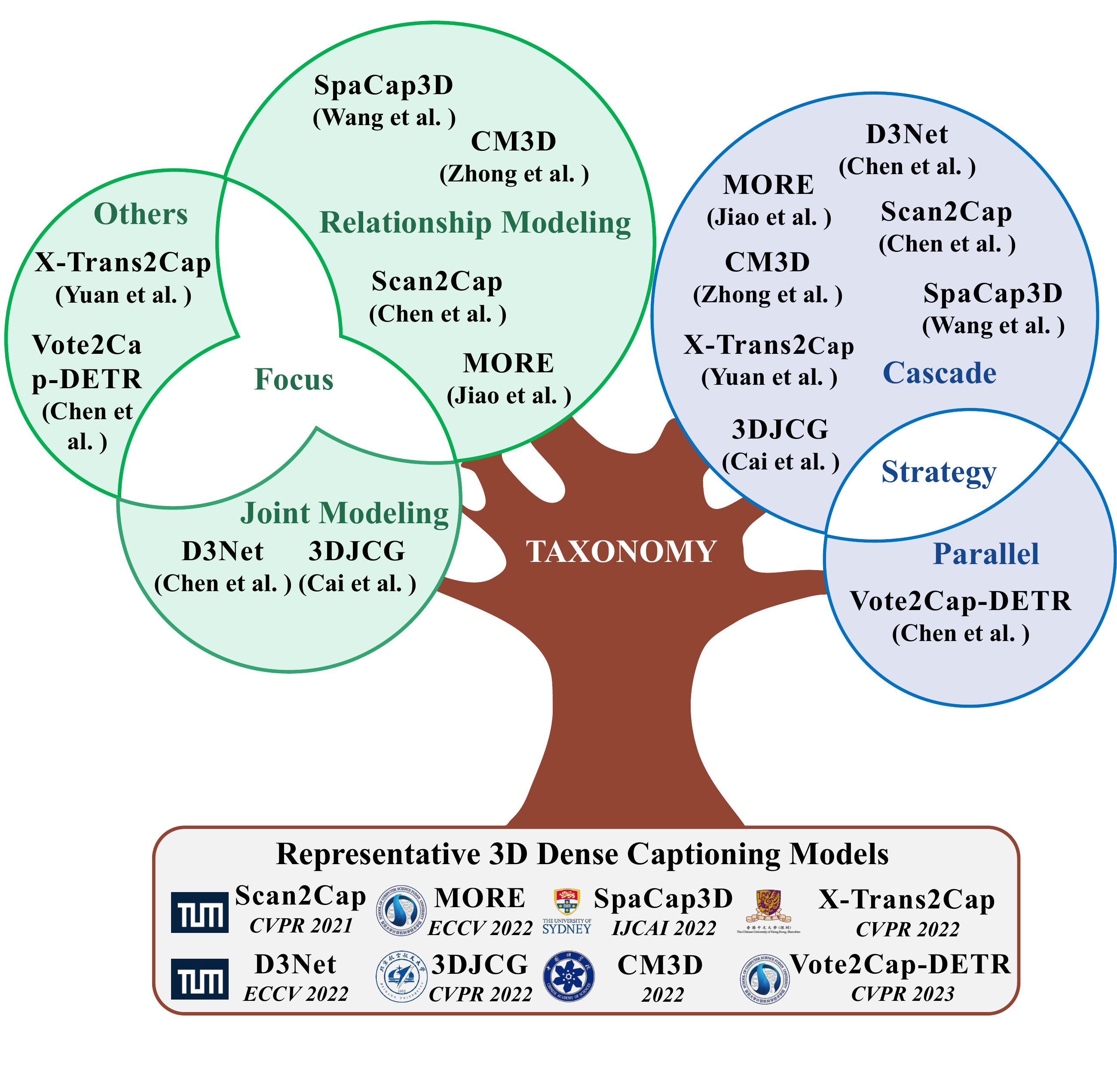}
  \caption{Classification of models in the context of 3D dense captioning. The existing approaches can be categorized based on their research focus and research strategy. With regards to the specific research motives and foci, most methods can be broadly classified into three groups: relationship modeling that focuses on building inter-object or context relationships; research on joint modeling that combines two distinct tasks; and other approaches that do not fall into these two categories. Regarding the research strategy employed to tackle the 3D dense captioning task, existing models can be categorized into the ``\emph{detection-then-captioning}" cascade strategy and the ``\emph{detection-and-captioning}" parallel strategy. }
\label{fig5}
\end{figure}

\subsection{Dense Video Captioning}
Building upon the concept of dense image captioning, Krishna \emph{et al.} introduced a more challenging task known as dense video captioning \cite{krishna2017dense}. Unlike traditional video captioning approaches \cite{lin2022swinbert,luo2020univl,seo2022endtoend}, which generate a single caption for an entire video, dense video captioning \cite{li2018jointly,mun2019streamlined,iashin2020multi,krishna2017dense} involves describing multiple events within a minute-long video. This task proves beneficial for the search and indexing of untrimmed, large-scale videos. 
The majority of dense video captioning techniques \cite{iashin2020better,iashin2020multi,wang2018bidirectional,wang2020event} follow a two-stage strategy. It begins with event localization \cite{gao2017turn,heilbron2016fast}, where events of interest within the video are identified, followed by event captioning \cite{gao2017video,lin2022swinbert} to generate descriptive captions for these localized events. These two-stage approaches bear a resemblance to the process employed in 3D dense captioning. 
To enhance the interaction between event localization and captioning, certain approaches \cite{li2018jointly,wang2021endtoend,zhou2018endtoend} propose performing these subtasks jointly. In particular, Li \emph{et al.} \cite{li2018jointly} employed a descriptiveness regression technique to dynamically adjust the temporal positions of individual event candidates and enable the integration of event proposal localization and event caption generation, resulting in a unified framework. Conversely, other works \cite{deng2021sketch, zhu2022endtoend} eliminate explicit event localization and instead ground each sentence in the video after generating a comprehensive video description paragraph.
With the emergence of multi-modal pre-trained models \cite{gan2020largescale,kim2021vilt,li2019unicodervl,luo2020univl}, Yang \emph{et al.} have recently explored integrating end-to-end dense video captioning into these large-scale pre-trained models \cite{yang2023vid2seq}, leveraging their capabilities for improved performance and scalability.

\subsection{3D Visual Grounding}
3D visual grounding focuses on localizing objects in 3D scenes based on their textual descriptions. It is closely related to 3D dense captioning but with key differences. Unlike 3D dense captioning generating descriptions for each object in a point cloud, 3D visual grounding aims to locate the object described by a specific input sentence. In analogy to human listening and speaking, 3D visual grounding simulates the process of listening, while 3D dense captioning is more akin to speaking, as described in \cite{takahashi2020d3net}. Dave \emph{et al.} \cite{chen2020scanrefer} introduced the dataset, the pioneer method, and the benchmarks for 3D visual grounding. Most contemporary 3D visual grounding methods consist of two phases: object detection \cite{sheng2023pdr,wang2023long} and description matching \cite{he2021transrefer3d,huang2021text,yang2021sat,yuan2021instancerefer}. In the object detection phase, objects were detected and segmented under the related 3D scenes. Subsequently, in the description matching phase, the textual descriptions were matched with the detected objects to identify the referred object in the scene. Early methods for 3D visual grounding relied on object detectors for scene localization. However, Chen \emph{et al.} \cite{takahashi2020d3net} proposed a novel hierarchical attention model that enables end-to-end training for both detection and grounding \cite{chen2022ham}, providing a unified and integrated pipeline. More recently, the technique of combining grounding tasks with dense captioning experience a growing trend \cite{cai20223djcg,takahashi2020d3net}, which explored the potential synergies between related tasks and opened up new research directions.

\section{Architecture}\label{sec:architecture}
\subsection{Task Definition}
3D dense captioning is a task that aims to achieve object-level 3D scene understanding by generating natural language descriptions for objects in 3D scenes through the analysis of 3D visual data \cite{chen2021scan2cap, wang2022spatiality, zhong2022contextual}. The input to the model is a 3D point cloud, which represents the geometric and appearance characteristics of objects in the scene, along with additional auxiliary features such as colors, normal vectors, and multi-view features. The point cloud data can be mathematically represented as a matrix $P \in \mathbb{R}^{ N \times F}$, where $N$ denotes the number of randomly sampled points per scene (typically 40,000), and $F$ represents the dimensionality of scene features, including point coordinates $(x, y, z)$, and other auxiliary features.
Most existing methods generate $M$ object proposals with $F'$-dimensional features prior to generating captions, where $M$ is typically set to a default value of 256. The text data, comprising the captions, is tokenized using the SpaCy library \cite{honnibal2017spacy} and represented as a matrix $W \in \mathbb{R}^{T \times 300}$ using GloVE word embeddings \cite{pennington2014glove}, where $T$ represents the number of tokens in the caption, and each word embedding vector has a dimension of 300.
During the training phase, the ground truth caption represented by $W$ is utilized to optimize the generated word token probabilities using a loss function (detailed discussion in Section \ref{sec:exam}). During the inference phase, only the point cloud data $P$ is fed into the model. The expected output is a set of bounding boxes with corresponding captions $(B_i, D_i)$, where $B_i$ denotes the bounding box coordinates and $D_i$ represents the generated description for each object.

\subsection{Main Framework}
The realm of 3D dense captioning has been predominantly influenced by the encoder-decoder architecture, which can be delineated into three principal components: scene encoder, relation module, and feature decoder, as depicted in Figure \ref{flowchart}. The scene encoder is responsible for extracting initial scene details, such as object-level visual features and contextual information from input point clouds with various 3D object detection methods, such as PointNet++ \cite{qi2017pointnet++}, VoteNet \cite{qi2019deep}, PointGroup \cite{jiang2020pointgroup}, 3DTER \cite{misra2021end}, and others \cite{sheng2023pdr,wang2023long}. The relationship module serves as a pivotal component in most 3D dense captioning models to model intricate connections within the scenes or cross-modal interactions. A well-designed relational module has been empirically shown to significantly enhance model performance, as substantiated by numerous ablation experiments. Feature decoder commonly employs sequential models such as GRU \cite{chung2014empirical} or Transformer \cite{vaswani2017attention} to further decode the attribute information and relation tokens, thereby generating pertinent captions as well as bounding boxes for the target objects. The stage of decoding is essential for generating precise and meaningful captions that accurately describe the objects in the 3D scenes.
\subsubsection{ Scene Encoder}
The scene encoder is tasked with extracting initial scene details. Previous research efforts \cite{chen2021scan2cap,jiao2022more,wang2022spatiality,yuan2022x,cai20223djcg,zhong2022contextual} have primarily utilized VoteNet or modified-VoteNet as the feature extraction backbone to obtain comprehensive visual features. The composition of the visual features may vary depending on the specific approach. For instance, CM3D \cite{zhong2022contextual} captures background environmental details, while X-Trans2Cap \cite{yuan2022x} focuses on object attributes. In recent approaches, more robust detectors such as PointGroup \cite{jiang2020pointgroup} and 3DETR \cite{misra2021end} have been employed as the feature extraction backbone. For example, D3Net \cite{chen2023end} adopts PointGroup with U-Net architecture \cite{ronneberger2015u}, while Vote2Cap-DETR \cite{chen2023end} utilizes a full transformer structure for 3DETR. These approaches have achieved notable performance in ablation studies, underscoring the significance of employing effective object detectors in the scene encoding stage. Additionally, most approaches\cite{chen2021scan2cap,jiao2022more,wang2022spatiality,yuan2022x,cai20223djcg,takahashi2020d3net,zhong2022contextual} also aggregate the object proposals and bounding boxes at this stage, while some \cite{chen2023end} handle this task in the decoding step, depending on the specific approach.

\subsubsection{Relation Module}
The relation module enables the modeling of intricate connections and cross-modal interactions within indoor scenes. The choice of relationship modeling approach depends on the specific requirements of the 3D dense captioning task and the characteristics of the scene understanding problem being addressed. There are several commonly used approaches for modeling relationships, including graph-based methods, transformer-based approaches, and knowledge distillation. Methods like Scan2Cap \cite{chen2021scan2cap}, MORE \cite{jiao2022more}, and D3Net \cite{takahashi2020d3net} utilize semantic scene graphs \cite{wang2019dynamic, feng2021free} to capture inter-object spatial location relationships. In these methods, object proposals are treated as nodes in the graph, and the relationships between objects are modeled as edges connecting the nodes. For instance, relationships such as top, front, left, or center can be learned between objects with a graph structure. These approaches also leverage the transitive property of relationships, where the relationship between two non-adjacent nodes can be inferred from a common node. This allows for reasoning about relationships between objects that are not directly connected in the graph.
3DJCG \cite{cai20223djcg} and SpaCap3D \cite{wang2022spatiality} employ transformer-based modules to build inter-object relationships. Specifically, 3DJCG utilizes a feature enhancement module comprising multi-head self-attention layers and fully connected layers to model inter-object relationships and enhance attribute characteristics. SpaCap3D uses a standard transformer encoder with a relation prediction head to capture object-to-object relations. These approaches leverage the self-attention mechanism of transformers to model relationships between objects and capture long-range dependencies in the scene. Knowledge distillation is another approach applied in some models, such as X-Trans2Cap \cite{yuan2022x}, which applies a knowledge distillation framework with a cross-modal fusion module to facilitate interaction between 3D object features and multiple modalities.

\subsubsection{Feature Decoder}
Feature decoder involves generating bounding boxes and captions for candidate objects, incorporating the attribute and relationship features learned in previous stages. However, in addition to Vote2Cap-DETR, most of the models only perform caption generation at this stage. Scan2Cap, MORE, and D3Net employ a GRU-based decoder with attention mechanisms to generate descriptions for candidate objects. Other methods, such as SpaCap3D, X-Trans2Cap, 3DJCG, and Vote2Cap-DETR, apply a transformer-based decoder for caption generation. Specifically, X-Trans2Cap utilizes a transformer decoder for both the teacher and student frameworks, directly incorporating the transformer architecture for caption generation \cite{yuan2022x}. SpaCap3D introduces an object-centric decoder with an improved masked self-attention mechanism \cite{wang2022spatiality}. 3DJCG utilizes a multi-head cross-attention network following a fully connected layer with a language prediction module as the caption head \cite{cai20223djcg}. Vote2Cap-DETR employs two identical standard transformer decoders, a position embedding, and a linear classification head for generating descriptions \cite{zhong2022contextual}. Notably, CM3D stands out by stacking multiple decoder layers for caption generation. This allows CM3D to capture richer contextual information for generating more detailed and coherent captions.
\begin{figure}
  \centering
  \includegraphics[width=\linewidth]{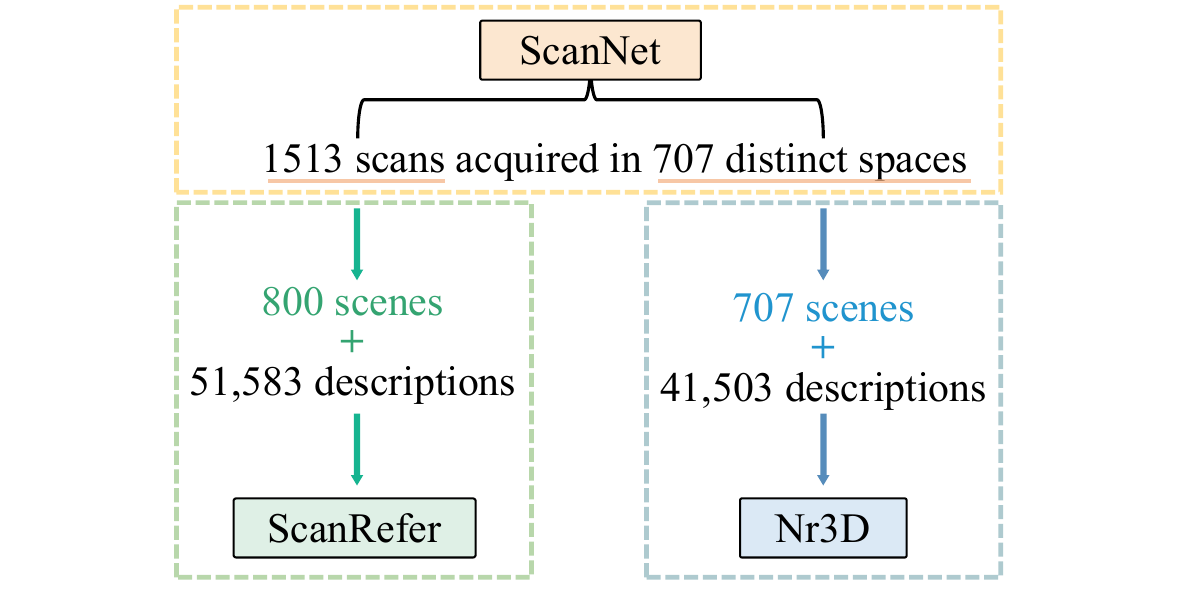}
  \caption{Illustration of relationship between ScanNet \cite{dai2017scannet}, ScanRefer \cite{chen2020scanrefer}, and Nr3D \cite{achlioptas2020referit3d}. ScanRefer and Nr3D are proposed with extra human-labeled descriptions based on the scenes in ScanNet.}
\label{fig1-2}
\end{figure}

\subsection{Model Classification}
The existing approaches in the field of 3D dense captioning can be categorized based on their research focus and research strategy. With regard to the specific research motives and foci, these methods can be broadly classified into three groups: relationship modeling, joint modeling, and other approaches that do not fall into these two categories. Regarding the research strategy employed to tackle the 3D dense captioning task, existing models can be classified into two categories: the ``\emph{detection-then-captioning}" cascade strategy and the ``\emph{detection-and-captioning}" parallel strategy. The classification of models with the context of 3D dense captioning is illustrated in Fig. \ref{fig5}.

\subsubsection{ Research Focus}
The research focus of 3D dense captioning has been explored from various perspectives, including research on relationship modeling, joint modeling, and other approaches.

\textbf{\itshape Relationship Modeling}. 
Scan2Cap is the pioneering method that introduces a graph-based attentive captioning framework to explore object relations. Building upon Scan2Cap, Jiao \emph{et al.} \cite{jiao2022more} propose an improved model called MORE, which incorporates multi-order relation mining based on graphs. MORE captures more complex inter-object relationships through progressive encoding. In contrast, ScanCap3D focuses specifically on capturing spatial relativity in 3D scenes using a spatiality-guided encoder-decoder transformer architecture. ScanCap3D achieves this while maintaining a simpler size and higher computational efficiency compared to Scan2Cap and MORE.
In contrast to previous works \cite{chen2021scan2cap,jiao2022more,wang2022spatiality} that mainly investigate inter-object interactions, neglecting contextual details, CM3D \cite{zhong2022contextual} addresses this limitation by incorporating rich contextual information. CM3D considers non-object details, background environments, and generates more detailed descriptions, thus providing a more comprehensive understanding of the scene.

\textbf{\itshape Joint Modeling}.
D3Net introduced unified networks for both 3D dense captioning and 3D visual grounding tasks in the context of 3D scenes. It originates from the same team that developed the Scan2Cap model for 3D dense captioning and utilized the ScanRefer dataset for 3D visual grounding \cite{chen2020scanrefer}. In D3Net, the overlapping aspects of these two tasks are integrated to create a unified model. This integration is motivated by the challenges posed by limited vision-language data, which can lead to overfitting, as well as the difficulty of generating unique descriptions for similar objects. To tackle these challenges, a joint speaker-listener architecture is introduced, where the ``speaker" corresponds to the captioning model and the ``listener" corresponds to the grounding task. This architecture enables semi-supervised training and facilitates the generation of distinctive descriptions \cite{takahashi2020d3net}.
Similarly, 3DJCG \cite{cai20223djcg} also explores the joint framework of 3D visual grounding and 3D dense captioning. The goal of this framework is to capture visual and textual information efficiently and effectively for generating comprehensive 3D captions \cite{cai20223djcg}. By identifying the complementary relationship between the two tasks, 3DJCG proposes an innovative transformer-based approach that incorporates three key components: a 3D object detector, a feature enhancement module, and a grounding or captioning head.

\begin{table}
 \caption{The statistics of the ScanRefer and Nr3D datasets. Both datasets comprise primarily 3D scenes, object labels, and manually annotated descriptions. However, the notable differences lie in their marking strategies. ScanRefer labels almost all objects in a scene, while Nr3D focuses on labeling specific categories that appear with higher frequency. }
  \label{tab5}
\renewcommand\arraystretch{1.3}
\tabcolsep0.3cm
\centering
\begin{tabular}{c|c|c}
\hline
Number \#  & ScanRefer & Nr3D   \\ \hline
Descriptions                 & 51,583    & 41,503 \\ \hline
Scenes                       & 800       & 707    \\ \hline
Objects / Contexts           & 11,046    & 5,878  \\ \hline
Object classes               & 265       & 76     \\ \hline
Objects / Contexts per scene & 13.81     & 8.31   \\ \hline
Descriptions per object      & 4.67      & 7      \\ \hline
Average length of descriptions         & 20.27     & 11.4   \\ \hline
\end{tabular}
\end{table}
\textbf{\itshape Others}.
To address the challenge of enriching 3D point clouds with 2D information while minimizing computational overhead, Yuan \emph{et al.} propose a transformer-based cross-modal teacher-student framework in X-Trans2Cap \cite{yuan2022x}. This framework utilizes knowledge distillation \cite{hinton2015distilling} to transfer rich appearance information, including texture awareness and color clues, from 2D images to 3D scenes. By employing the teacher-student paradigm, X-Trans2Cap effectively integrates 2D information into the 3D captioning process without imposing a significant computational burden \cite{yuan2022x}. In a more recent development, Vote2Cap-DETR adopts a fully end-to-end transformer encoder-decoder architecture to enable parallel detection and captioning \cite{chen2023end}. This approach eliminates the need for separate detection and captioning stages, resulting in a more efficient and integrated processing framework.

\subsubsection{Research Strategy}
Regarding the research strategy employed to tackle the 3D dense captioning task, existing models can be categorized into two main categories. The first category is a cascade strategy, where object detection is followed by captioning, namely the ``\emph{detection-then-captioning}" cascade strategy. The second category is a parallel strategy, where object detection and captioning are addressed simultaneously, namely the ``\emph{detection-and-captioning}" parallel strategy. 
\begin{figure*}[h]
  \centering
  \includegraphics[width=\linewidth]{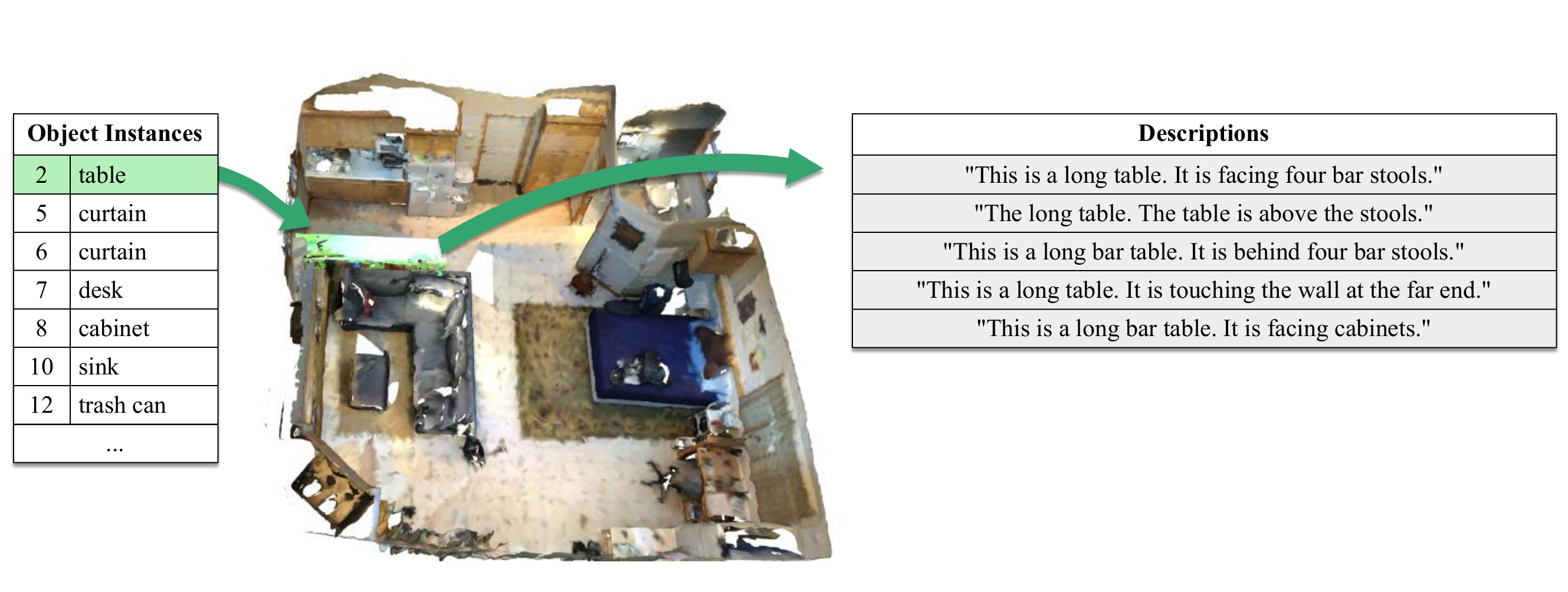}
 \caption{Illustration of a representative example of scene 0000\_00 from the ScanRefer Dataset \cite{chen2020scanrefer}. The ScanRefer dataset is known for its comprehensive object labeling approach, wherein nearly every object in the scene is annotated, accompanied by five corresponding descriptions.}
\label{fig1-3}
\end{figure*}

\begin{table}[h!]
 \caption{The statistics of the standard split of ScanRefer dataset.}
\label{tab2}
\renewcommand\arraystretch{1.3}
\tabcolsep0.3cm
\begin{tabular}{ccccc}
\hline
Number \#                                 & Train  & Val   & Test  & Total  \\ \hline
 Descriptions            & 36,665 & 9,508 & 5,410 & 51,583 \\
 Scenes                  & 562    & 141   & 97    & 800    \\
 Objects                 & 7,875  & 2,068 & 1,103 & 11,046 \\
 Objects per scene       & 14.01  & 14.67 & 11.37 & 14.14  \\
 Descriptions per scene  & 65.24  & 67.43 & 55.77 & 65.68  \\
 Descriptions per object & 4.66   & 4.60  & 4.90  & 4.64   \\ \hline
\end{tabular}
\end{table}
\begin{figure}[h]
  \centering
  \includegraphics[width=\linewidth]{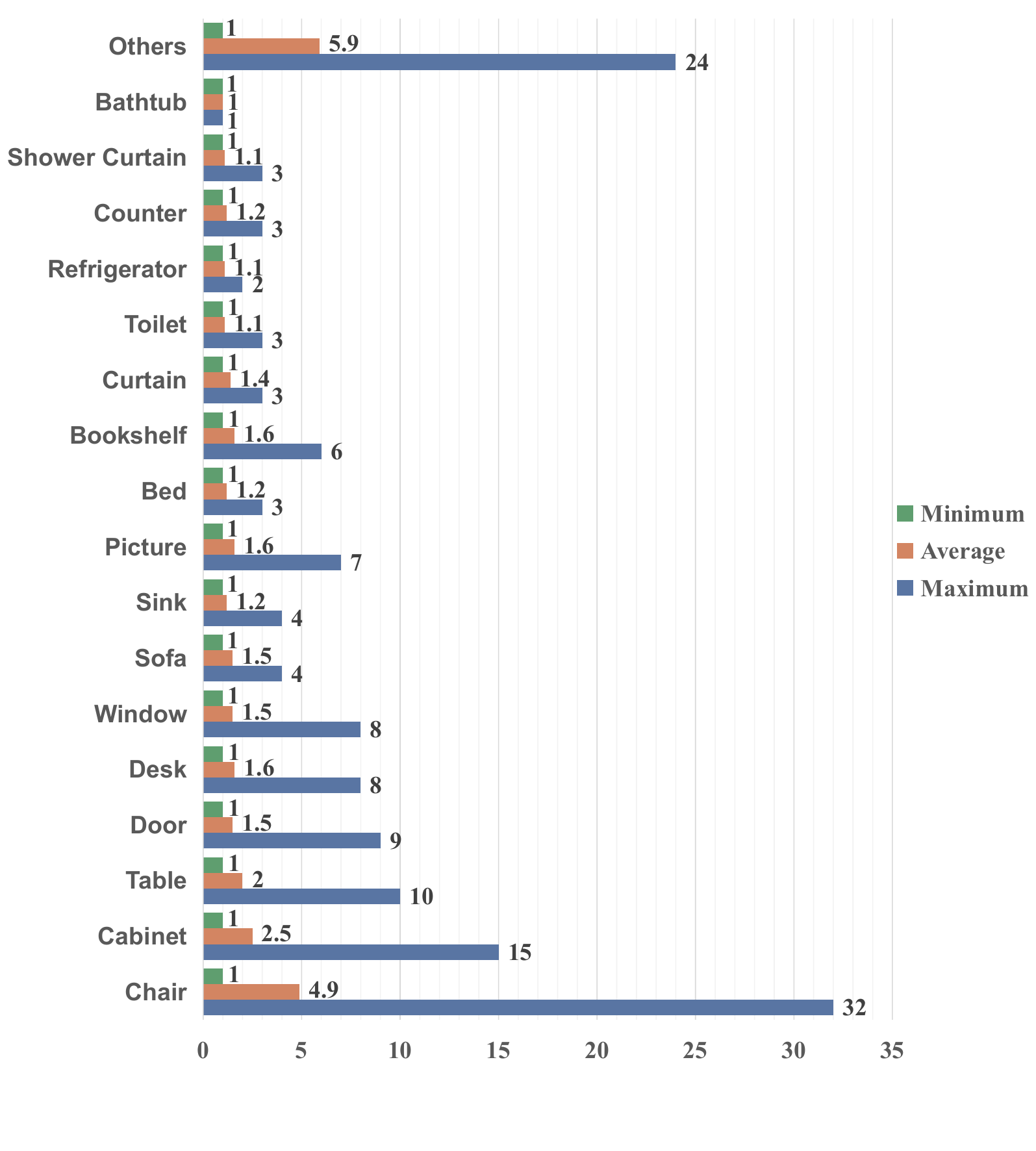}
  \caption{Illustration of the manifestation of 18 broad categories comprised of 256 sub-categories in a 3D indoor scene, primarily encompassing furniture items such as chairs, tables, and cabinets. ``Maximum", ``Average", and ``Minimum" indicate the maximum, average, and minimum number for a certain type of object that appears in a scene, respectively. For example, there are up to 32 chairs in a scene.}
  \label{fig1-5}
\end{figure}

\textbf{\itshape Cascade Strategy}. In the case of the cascade strategy, most methods follow a ``detect-then-caption" paradigm. The model first generates a set of candidate objects with their bounding boxes and then performs feature enhancement and relationship inference around these candidate objects. Subsequently, corresponding descriptions are generated based on these object-centric features. However, this cascade strategy is not without its limitations \cite{chen2023end}. On the one hand, the performance of captioning is heavily influenced by the accuracy of the detection results. On the other hand, the hand-crafted components designed in previous detectors may result in poor performance.

\textbf{\itshape Parallel Strategy}. To overcome these limitations, Vote2Cap-DETR takes a different approach by employing a parallel strategy. They reverse the order of captioning and detection by utilizing 3DETR \cite{misra2021end}, a successful transformer-based architecture for 3D object detection, as a feature encoder for generating scene tokens. Spatial bias and content-aware features are introduced to refine the initial object queries into more precise vote queries. Subsequently, two independent decoder heads are designed for object location and caption generation simultaneously. This one-step attention-driven strategy helps to mitigate over-reliance on detectors and reduce the hard-coded limitations of object detectors. Additionally, it derives more effective localization and more accurate captioning.
\begin{figure*}[h]
  \centering
  \includegraphics[width=\linewidth]{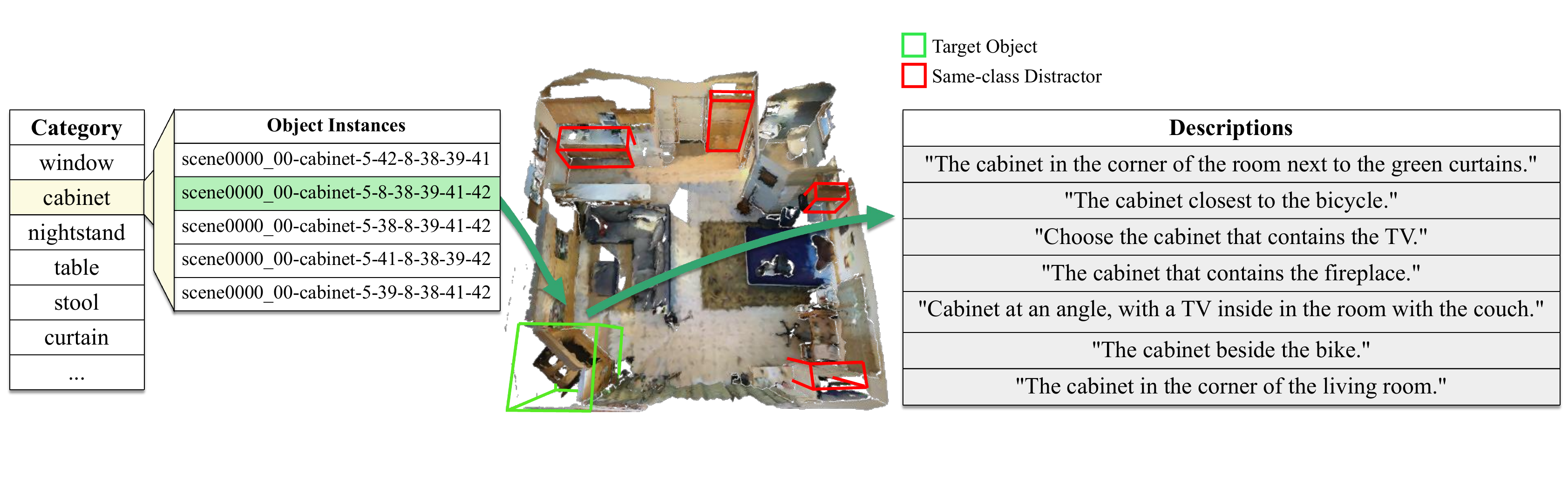}
  \caption{Illustration of a typical example of  scene 0000\_00 from Nr3D dataset \cite{achlioptas2020referit3d}. The Nr3D dataset performs selective annotation of commonly occurring object categories, mainly encompassing multiple objects, with each object being associated with seven distinct descriptions.}
\label{fig1-4}
\end{figure*}

\section{Datasets and Evaluation Metrics}\label{sec:task}
\subsection{Datasets Analysis}
The availability of large-scale data is crucial for achieving superior performance in vision and language tasks. However, capturing and annotating 3D data pose significant challenges, resulting in smaller RGB-D datasets \cite{shotton2013scene,xiao2013sun3d,song2016deep,hua2016scenenn} for 3D scenes compared to their 2D counterparts. As a result, researchers resort to utilizing manufactured data to compensate for the absence of real-world data \cite{wu20153d,chang2015shapenet}. To address this limitation, Dai \emph{et al.} introduced the ScanNet dataset \cite{dai2017scannet}, which is a richly-annotated 3D indoor scan dataset of real-world environments captured employing a self-created RGB-D capture system. 

The ScanNet dataset comprises 2.5 million views obtained from 1513 scans, which are obtained from 707 distinct spaces, making it larger in scale compared to other popular datasets \cite{silberman2012indoor,sturm2012benchmark,xiao2013sun3d,savva2016pigraphs,dai2017scannet}, such as NYUv2 \cite{silberman2012indoor}, SUN3D \cite{xiao2013sun3d}, and SceneNN \cite{hua2016scenenn}. However, it should be noted that ScanNet is primarily designed for tasks such as 3D object classification, semantic voxel labeling, and 3D object retrieval with instance-level category annotation in a 3D point cloud and may not be directly suitable for 3D dense captioning and 3D visual grounding tasks.
To overcome this limitation, ScanRefer \cite{chen2020scanrefer} and Nr3D \cite{achlioptas2020referit3d} were developed as datasets specifically tailored for 3D dense captioning, building on top of ScanNet. The relationship between these datasets is depicted in Fig. \ref{fig1-2}. Both ScanRefer and Nr3D select partial point cloud scenes from ScanNet and provide additional human-labeled descriptions for the objects in each scene. Furthermore, they also offer online browsing sites for visualizing the data, making them valuable resources for developing and evaluating models in the field of 3D dense captioning and 3D grounding tasks.

\textbf{\itshape ScanRefer}. 
ScanRefer is widely recognized as a prominent dataset for 3D dense captioning tasks, offering a substantial number of natural language descriptions for objects in the scans from the ScanNet dataset \cite{chen2020scanrefer}.  Statistically, the dataset encompasses 51,583 detailed and diverse descriptions for 11,046 objects in 800 ScanNet scans, covering over 250 types of common indoor objects and including attributes such as color, size, shape, and spatial relationships. Approximately five manual annotations are provided for each object in each scene, resulting in a rich and comprehensive dataset. The detailed statistics of the ScanRefer are presented in the second column of Table \ref{tab5}. An example of a typical scene from ScanRefer, specifically scene 0000\_00, is illustrated in Fig. \ref{fig1-3}. Following the official split of ScanNet \cite{dai2017scannet}, the dataset is partitioned into training, validation, and test sets quantitively with 36,665, 9,508, and 5,410 samples, respectively. Since the unseeable testing split has not been officially released, most experiments are performed on the validation set \cite{chen2021scan2cap}. The distribution statistics of the ScanRefer dataset are provided in Table \ref{tab2}. Additionally, Fig. \ref{fig1-5} presents the frequency of several major object categories in ScanRefer.

\textbf{\itshape Nr3D}. 
The 3D Natural Reference (Nr3D) dataset is one of the datasets from ReferIt3D \cite{achlioptas2020referit3d}, along with the synthetic language descriptions dataset Sr3D \cite{achlioptas2020referit3d}. Fig. \ref{fig1-4} presents a typical example of scene 0000\_00 from Nr3D. Nr3D focuses on fine-grained objects in 3D space, which makes it more challenging compared to ScanRefer as the reference object is of the same kind \cite{yuan2022x}. Nr3D consists of 41,503 natural, free-form utterances describing objects belonging to one of 76 fine-grained object classes within 5,878 communication contexts. The communication contexts are expressed as unique sets denoted as $\{S, C\}$, where $S$ represents one of the 707 distinct ScanNet scenes, and $C$ denotes the fine-grained class of $S$. In other words, a context represents the same object type in a scene. Similar to ScanRefer, Nr3D uses the official ScanNet splits for experiments. The statistics of Nr3D are presented in the right column of Table \ref{tab5}.
\begin{table*}[h!]
\caption{Evaluation Metrics for 3D Dense Captioning. The table presents a comprehensive overview of various evaluation metrics employed for assessing the performance of 3D dense captioning models. Notably, CIDEr, being a metric that closely aligns with human assessments, is considered the most significant among the listed metrics.}
\label{tab1}
\renewcommand\arraystretch{1.3}
\tabcolsep0.3cm
\centering
\begin{tabular}{c|c|c|c|c|c}
\hline
Metric & Acronym & Based On                                                            & Original Task              & Advantages                                                             & Drawbacks                                                                            \\ \hline
BLEU-4 & B-4     & Precision                                                           & Machine translation        & Simple and efficient                                                   & \begin{tabular}[c]{@{}c@{}}Grammar issues and \\ limited by text length\end{tabular} \\ \hline
ROUGE  & R       & Recall                                                              & Machine text summarization & Simple and orderly                                                     & Disregard synonyms                                                                   \\ \hline
METEOR & M       & \begin{tabular}[c]{@{}c@{}}Precisiond \\ Recall \\ Pen\end{tabular} & Machine translation        & \begin{tabular}[c]{@{}c@{}}Consider stems \\ and synonyms\end{tabular} & \begin{tabular}[c]{@{}c@{}}Reliance on external \\ knowledge sources\end{tabular}    \\ \hline
CIDEr  & C       & TF-IDF                                                              & Image captioning           & Better human assessments                                               & Hard to optimize                                                                     \\ \hline
\end{tabular}
\end{table*}

\subsection{Evalution Metrics}
To rigorously assess the effectiveness of various methods for the 3D dense captioning task, performance evaluation is typically conducted from both captioning and detection perspectives, following established standards in the field. Captioning performance is evaluated using widely used text generation metrics, including CIDEr \cite{vedantam2015cider}, BLEU-4 \cite{papineni2002bleu}, METEOR \cite{banerjee2005meteor}, and ROUGE \cite{lin2004rouge1}, denoted as $C$, $B$-4, $M$, and $R$ respectively. Additionally, Chen \emph{et al.}  proposed a novel evaluation metric \cite{chen2021scan2cap} to calculate Intersection-over-Union (IoU) scores between the ground-truth bounding boxes and the predicted bounding boxes, formulated as follows:
\begin{equation}
      E@kIoU = \frac{1}{N} {\textstyle \sum_{i= 0}^{N}} E_{i} U_{i}\ ; \  U_{i}= \left\{\begin{matrix}1,\,IoU\ge k
 \\0,\,IoU< k

\end{matrix}\right.
\nonumber 
\end{equation}
where $E$ denotes the specific evaluation metrics, including CIDEr, BLEU-4, METEOR, and ROUGE (denoted as $C$, $B-4$, $M$, and $R$, respectively). The number of detected object bounding boxes is denoted as $N$. The value of $U_i$ is utilized to determine the IoU threshold, and $k$ is commonly set to 0.5 or 0.25. Additionally, the object detection metric employs the Mean Average Precision (mAP) thresholded by IoU, providing a comprehensive assessment of the detection performance in localizing objects.

In the past, image captioning evaluation heavily relied on machine translation evaluation metrics such as BLEU \cite{papineni2002bleu}, ROUGE \cite{lin-2004-rouge}, and METEOR \cite{banerjee2005meteor}, which lack relevance to the human assessment. More recently, with the introduction of CIDEr \cite{vedantam2015cider} and SPICE \cite{anderson2016spice} indicators designed explicitly for image captioning, this issue has been addressed. In 3D dense captioning, the metrics BLEU, ROUGE, METEOR, and CIDEr are selected for captioning assessment. We provide a detailed introduction to these four metrics below and organize them in Table \ref{tab1}.

\subsubsection{BLEU}
Bilingual Evaluation Understudy (BLEU) is a well-established metric for evaluating machine translation outputs based on precision. It involves matching different parts of a candidate text with reference texts and then calculating the percentage of successfully matched sequences. However, the basic calculation of BLEU is limited to unigram precision, which matches individual words. Many modern approaches adopt modified n-gram precision, which considers sequences of n consecutive words. For instance, in the context of 3D dense captioning, BLEU-4 is commonly used, which considers four consecutive words in the generated text as one matching unit.
One limitation of BLEU is that it may not be able to detect syntactic issues, as successful sequences of matches may appear in the wrong order. This can result in inaccurate scores. Additionally, shorter generated texts may have higher chances of being incomplete, leading to unreliable BLEU scores \cite{callison2006koehn,lin2004automatic}. Therefore, BLEU tends to favor candidate texts that have a similar length to the reference captions to mitigate this issue.

\subsubsection{ROUGE} 
Recall-Oriented Understudy for Gisting Evaluation (ROUGE) is a generally employed metric for assessing the quality of text summarization generated by machine learning algorithms. It is based on recall and measures the similarity between the candidate and reference summaries with various n-gram-based and sequence-based statistics. ROUGE has several variants, including ROUGE-N, ROUGE-W, ROUGE-L, and ROUGE-S, built based on different considerations.
ROUGE-N estimates the n-gram recall between the candidate and reference summaries, where n denotes the length of the n-gram. ROUGE-W considers the weighted longest common subsequence, considering the essence of words in the sequences. ROUGE-L focuses on calculating the sentence-level structure similarity, performing the statistics of the longest common subsequence and longest co-occurrence in the n-grams sequence. ROUGE-S incorporates skip-bigram co-occurrence statistics, allowing for measuring the similarity of non-consecutive words.
It is worth mentioning that although ROUGE is widely adopted for evaluating text summarization tasks, it may not perform well in the context of multi-document text summaries. Thus, it is essential to carefully consider the appropriate variant of ROUGE for the specific task being evaluated.

\subsubsection{METEOR}
Metric for Evaluation of Translation with Explicit ORdering (METEOR ) is employed to measure the machine-translated language. The metric is based on the harmonic mean of unigram precision and recall. Moreover, it not only matches candidate sentences with standard reference captions but is also equipped with stemming and synonym-matching capabilities. To a certain extent, METEOR makes up for the shortcomings of the above two indicators, reflecting grammar and sentence fluency. However, it relies heavily on external knowledge sources and only considers unigram, which may not capture the nuances of higher-order language structures or semantic meanings.

\subsubsection{CIDEr}
In contrast to previous metrics, which often lack correlation with human consensus, CIDEr incorporates Term-Frequency Inverse Document Frequency (TF-IDF) weighting for each n-gram, resulting in more accurate human assessments \cite{robertson2004understanding}. Notably, CIDEr disregards n-grams that do not appear in the reference sentence and instead encodes the frequency of n-gram occurrences in the reference statement. N-grams that occur more frequently are assigned lower weights, as they are less likely to contain essential information, such as common phrases like \emph{``this is a''}. As a result, CIDEr has been deemed as a pivotal metric for evaluating the quality of 3D dense captioning.
Despite its significance, CIDEr does possess a limitation in terms of optimization. The optimization process for CIDEr can be challenging due to its intricate weighting schemes based on TF-IDF, and the scoring may not always align perfectly with human judgments.

\section{Experimental Details}\label{sec:exam}
\subsection{Loss Function}
Most models incorporate three key components, namely detection loss, caption loss, and relative direction loss, in their loss functions. Although we use the same notation for convenience, it is crucial to recognize that these components are not necessarily identical. Each model customizes its loss function to suit its unique characteristics. For example, the 3DJCG model specifically emphasizes enhancing the detection loss. Further details regarding the adaptations of each model are provided below.
In terms of training techniques, Maximum Likelihood Estimation (MLE) and Self-Critical Sequence Training (SCST) \cite{rennie2017self} are the primary categories. Models such as Scan2Cap, MORE, SpaCap3D, and 3DJCG follow the MLE training scheme, while other methods like X-Trans2Cap, D3Net, CM3D, and Vote2Cap-DETR integrate both MLE and SCST schemes. When comparing these two techniques, SCST consistently outperforms pure MLE. SCST leverages the output of test-time inference to normalize the rewards it receives, eliminating the need for reward signal estimation and normalization. These findings have been further supported by ablation experiments conducted with select methods \cite{chen2023end,takahashi2020d3net}.

\subsubsection{Scan2Cap}
The loss function employed in Scan2Cap comprises three distinct components, namely, the detection loss denoted as $\mathcal{L}_{\mathrm{det}}$, the relative orientation loss denoted as $\mathcal{L}_{\mathrm{rela}}$, and the caption loss denoted as $\mathcal{L}_{\mathrm{cap}}$. Specifically, the detection loss $\mathcal{L}_{\mathrm{det}}$ is formulated based on the approach proposed by Qi \emph{et al.} \cite{qi2019deep}, which encompasses four individual loss terms: vote regression loss, objectness binary classification loss, box regression loss, and semantic classification loss, as detailed in \cite{qi2019deep}. The relative orientation loss $\mathcal{L}_{\mathrm{rela}}$ is computed using a cross-entropy loss function, following the methodology employed in previous works \cite{xu2015show,vinyals2015show}. The caption loss $\mathcal{L}_{\mathrm{cap}}$ is determined using a conventional cross-entropy loss.
\begin{equation}
\setlength{\abovedisplayskip}{10pt}
\setlength{\belowdisplayskip}{10pt}
 \begin{aligned}
 &\mathcal{L}=\alpha\mathcal{L}_{\mathrm{det}}+ \beta\mathcal{L}_{\mathrm{rela}}+ \gamma\mathcal{L}_{\mathrm{cap}}\\
 &\mathcal{L}_{\mathrm{det}}= \theta _{1} \mathcal{L}_{\mathrm{vote}}+\theta _{2}  \mathcal{L}_{\mathrm{obj}} + \theta _{3} \mathcal{L}_{\mathrm{box}}+\theta _{4}  \mathcal{L}_{\mathrm{sem}}
 \end{aligned}
\end{equation}
where the hyperparameters are set as $\alpha=10$, $\beta=1$, $\gamma=0.1$, $\theta _{1}=1$, $\theta _{2}=0.5$, $\theta _{3}=1$, $\theta _{4}=0.1$, respectively.

\subsubsection{ MORE}
The loss function utilized in MORE is similar to that of Scan2Cap, with the exception of the relative orientation loss $\mathcal{L}_{\mathrm{rela}}$ component.
\begin{equation}
 \mathcal{L}=\alpha \mathcal{L}_{\mathrm{det}}+\beta \mathcal{L}_{\mathrm{cap}}
\end{equation}
where the hyperparameters $\alpha$ and $\beta$ are set to specific values, i.e., $\alpha=10$ and $\beta=0.1$, respectively.

\subsubsection{X-Trans2Cap} 
The loss function of X-Trans2Cap comprises two main components: the feature alignment loss, denoted as $\mathcal{L}_{\mathrm{align}}$, and the captioning loss, denoted as $\mathcal{L}_{\mathrm{cap}}$. The feature alignment loss $\mathcal{L}_{\mathrm{align}}$ is based on a Smooth-L1 regression loss, while the captioning loss $\mathcal{L}_{\mathrm{cap}}$ is a weighted combination of a cross-entropy loss function denoted as $\mathcal{L}_{\mathrm{cro}}$ and a reward based on the CIDEr-D score \cite{anderson2018bottom} denoted as $\mathcal{L}_{\mathrm{CIDEr}}$.
\begin{equation}
 \begin{aligned}
 &\mathcal{L}= \alpha \mathcal{L}_{\mathrm{align}}+ \mathcal{L}_{\mathrm{cap}}\\
 &\mathcal{L}_{\mathrm{cap}}= \beta \mathcal{L}_{\mathrm{cro}}+ \gamma \mathcal{L}_{\mathrm{CIDEr}}
 \end{aligned}
\end{equation}
where the hyperparameters are set as $\alpha=1$, $\beta=1$ and $\gamma=0.1$, respectively.

\subsubsection{SpaCap3D}
The loss function of SpaCap3D is mathematically represented in Eq. \ref{eq:loss4}, where the relational loss denoted as $\mathcal{L}_{\mathrm{rela}}$ is constructed on top of a three-class cross-entropy loss, which serves as a guide for spatial relation learning. The components of $\mathcal{L}_{\mathrm{det}}$ and $\mathcal{L}_{\mathrm{cap}}$ in SpaCap3D closely align with the approach employed in Scan2Cap.
\begin{equation}\label{eq:loss4}
 \mathcal{L}=\alpha \mathcal{L}_{\mathrm{det}}+ \beta \mathcal{L}_{\mathrm{rela}}+\gamma \mathcal{L}_{\mathrm{cap}}
\end{equation}
where the hyperparameters $\alpha$, $\beta$, and $\gamma$ are set to specific values, i.e., $\alpha=10$, $\beta=1$ and $\gamma=0.1$, respectively.

\subsubsection{3DJCG}
The loss function of 3DJCG comprises both captioning and grounding components, as it is a unified task. The formulation of the loss function is outlined in Eq. \ref{eq:loss5}. Notably, the grounding loss and captioning loss employ similar loss functions, with the former utilizing an average cross-entropy loss while the latter utilizes a conventional cross-entropy loss. It is worth mentioning that the detection loss has been enhanced following approach \cite{qi2019deep}, and the bounding box loss has been replaced by the boundary regression loss  \cite{tian2019fcos} denoted as $\mathcal{L}_{\mathrm{bbox\text{-}reg}}$.
\begin{equation}\label{eq:loss5}
 \begin{aligned}
 &\mathcal{L}= \alpha\mathcal{L}_{\mathrm{det}}+ \beta\mathcal{L}_{\mathrm{gro}}+\gamma\mathcal{L}_{\mathrm{cap}}\\
 &\mathcal{L}_{\mathrm{det}}= \theta _{1} \mathcal{L}_{\mathrm{vote}}+ \theta _{2} \mathcal{L}_{\mathrm{obj}} + \theta _{3} \mathcal{L}_{\mathrm{bbox\text{-}reg}}+ \theta _{4} \mathcal{L}_{\mathrm{sem}}
 \end{aligned}
\end{equation}
where the hyperparameters are set as $\alpha=1$, $\beta=0.3$, $\gamma=0.2$, $\theta _{1}=10$, $\theta _{2}=1$, $\theta _{3}=200$ and $\theta _{4}=1$, respectively.

\subsubsection{D3Net}
The comprehensive loss function employed in D3Net is presented in Eq. \ref{eq:loss6}. The detection loss $\mathcal{L}_{\mathrm{det}}$ is introduced from PointGroup \cite{jiang2020pointgroup}. It encompasses multiple components, namely, cross-entropy loss, L1 regression loss, the means of minus cosine similarities, and binary cross-entropy loss, each contributing to the overall loss. Similarly, $\mathcal{L}_{\mathrm{rela}}$ is fully adopted from Scan2Cap as the relative orientation loss for guiding spatial relation learning.
The joint loss of grounding and captioning, denoted as $\mathcal{L}_{\mathrm{gro}\text{-}\mathrm{cap}}$, represents the most intricate and complex aspect of D3Net's loss function, which is trained using reinforcement learning techniques \cite{rennie2017self,luo2018discriminability}. It comprises three key components: a maximize reward function $R(\hat{C}, I)$, a baseline reward function $R\left(C^{*}, I\right)$, and a caption loss $\mathcal{L}_{\mathrm{cap}}$. Specifically, the reward function $R(\hat{C}, I)$ is a weighted sum of three terms: the CIDEr score of the sampled captioning $R^{'} (\hat{C}, I)$, the localization loss $\mathcal{L}_{\mathrm{loc}}$, and the language object classification loss $\mathcal{L}_{\mathrm{locl}}$, with the last two being closely related to the grounding task and both employing the cross-entropy loss function. Meanwhile, the loss $\mathcal{L}_{\mathrm{cap}}$ is a standard captioning loss derived from the MLE strategy.
\begin{equation}\label{eq:loss6}
 \begin{aligned}
 &\mathcal{L}= \mathcal{L}_{\mathrm{det}}+ \mathcal{L}_{\mathrm{gro\text{-}cap}}+\alpha\mathcal{L}_{\mathrm{rela}}\\
&\mathcal{L}_{\mathrm{det}}= \mathcal{L}_{\mathrm{sem}}+ \mathcal{L}_{\mathrm{obj\text{-}reg}}+\mathcal{L}_{\mathrm{obj\text{-}dir}}+ \mathcal{L}_{\mathrm{c\text{-}score}}\\
&\mathcal{L}_{\mathrm{gro}\text{-}\mathrm{cap}}(\theta) \approx\left(R(\hat{C}, I)-R\left(C^{*}, I\right)\right)\mathcal{L}_{\mathrm{cap}}\\
&\mathcal{L}_{\mathrm{cap}}(\theta)=-\sum_{t=1}^{T} \log p\left(\hat{c_{t}} \mid \hat{c_{1}}, \ldots, \hat{c}_{t-1} ; I, \theta\right)\\
&R(\hat{C}, I)=R^{'} (\hat{C}, I)-\beta\left[\mathcal{L}_{\mathrm{loc}}(\hat{C})+\gamma\mathcal{L}_{\mathrm{locl}}(\hat{C}) \right ]
 \end{aligned}
\end{equation}
where the hyperparameters $\alpha$, $\beta$, and $\gamma$ are set to specific values, i.e., $\alpha$=0.3, $\beta$=0.1, and $\gamma$=1, respectively. The generated token at time step $t$ is denoted as $\hat{c}_{t}$, and the visual signal is represented by $I$. The model parameters are denoted as $\theta$, and the generated description is denoted as $\hat{C}$, which can be expressed as $\hat{C}=\left \{\hat{c}_{1},\ldots,\hat{c}_{t}  \right \}$.

\subsubsection{CM3D}
CM3D combines MLE training objectives with a SCST, which is formulated as follows.
\begin{equation}\label{eq:loss7}
\begin{aligned}
&\mathcal{L}_{mle}=-\sum_{i=1}^{T} \log \hat{P}\left(c_{i}^{t+1} \mid \mathcal{V}, c_{i}^{[1: t]}\right)\\
&\mathcal{L}_{scst}=-\sum_{i=1}^{k}\left(R\left(\hat{c}_{i}\right)-R(\hat{g})\right) \cdot \frac{1}{\left|\hat{c}_{i}\right|} \log \hat{P}\left(\hat{c}_{i} \mid \mathcal{V}\right)
\end{aligned}
\end{equation}
where $c_{i}^{t+1}$ represents the $(t+1)^{th}$ word, and the first $t$ words are denoted as $c_{i}^{[1: t]}$. The visual clue is represented as $\mathcal{V}$. The function $R\left(\cdot\right)$ denotes the CIDEr reward function, while $\hat{c}_{i}$ represents the generated multiple captions $c_{i}^{1}, \ldots, c_{i}^{k}$ with a beam size of k, and $\hat{g}$ serves as a baseline.

\subsubsection{Vote2Cap-DETR}
The loss function of Vote2Cap-DETR comprises three components: vote query loss $\mathcal{L}_{\mathrm{vq}}$, detection loss $\mathcal{L}_{\mathrm{det}}$, and caption loss $\mathcal{L}_{\mathrm{cap}}$. The vote query loss $\mathcal{L}_{\mathrm{vq}}$ is adopted from VoteNet, while the caption loss is fine-tuned with the self-critical sequence training strategy after maximum likelihood estimation. As for the detection loss $\mathcal{L}_{\mathrm{det}}$, it updates the weights on top of the original 3DETR loss to refine the model's performance in object detection. Specifically, $\mathcal{L}_{\mathrm{det}}$ consists of four parts $\mathcal{L}_{giou}$, $\mathcal{L}_{cls}$, $\mathcal{L}_{center\text{-}reg}$, and $\mathcal{L}_{size\text{-}reg}$, indicating the $gIoU$ loss, the box size classification loss, the box center regressing loss, and the box size regressing loss, respectively.
\begin{equation}\label{eq:loss8}
\setlength{\belowdisplayskip}{10pt}
\begin{aligned}
&\mathcal{L}=\alpha \mathcal{L}_{v q}+\beta \sum_{i=1}^{n_{\text {dec-layer }}} \mathcal{L}_{\text {det }}+\gamma \mathcal{L}_{\text {cap }}\\
&\mathcal{L}_{det}=\theta _{1} \mathcal{L}_{giou}+ \theta _{2} \mathcal{L}_{cls} + \theta _{3}  \mathcal{L}_{center\text{-}reg}+ \theta _{4} \mathcal{L}_{size\text{-}reg}
\end{aligned}
\end{equation}
where the hyperparameters are set as $\alpha=10$, $\beta=1$, $\gamma=5$, $\theta _{1}=10$, $\theta _{2}=1$, $\theta _{3}=5$, $\theta _{4}=1$, respectively.

\begin{table*}[h!]
  \caption{Experimental results of different models on the ScanRefer Validation Set. The best-reported result from the original paper is used as the benchmark standard. ``-" signifies that neither the original paper nor any subsequent works provide such results. The term ``3D" refers to methods that solely utilize coordinate information, while ``2D+3D" incorporates additional auxiliary features such as colors, normals, multi-view features, etc., along with point coordinates. ``GA" denotes the use of a GRU (Gated Recurrent Unit) backbone with an attention mechanism, and ``KD" stands for Knowledge Distillation. Caption metrics, including CIDEr, BLEU-4, METEOR, and ROUGE (denoted as C, B-4, M, and R, respectively), are adopted to evaluate the quality of captions, and mAP (mean Average Precision) is employed as the detection metric.}
	\renewcommand\arraystretch{1.4}
	\tabcolsep0.3cm
	\centering
  \label{tab3}
\begin{tabular}{c|l|c|ccccc|cccc}
\hline
\multirow{2}{*}{Method}        & \multicolumn{1}{c|}{\multirow{2}{*}{Architecture}} & \multirow{2}{*}{Data} & \multicolumn{5}{c|}{IoU=0.5}                                              & \multicolumn{4}{c}{IoU=0.25}                                      \\
                               & \multicolumn{1}{c|}{}                              &                       & C              & B-4            & M              & R              & mAP   & C              & B-4            & M              & R              \\ \hline
\multirow{2}{*}{Scan2Cap}      & \multirow{2}{*}{VoteNet  Graph GA}        & 3D                    & 35.20          & 22.36          & 21.44          & 43.57          & 29.13 & 53.73          & 34.25          & 26.14          & 54.95          \\
                               &                                                    & 2D+3D                 & 39.08          & 23.32          & 21.97          & 44.78          & 32.21 & 56.82          & 34.18          & 26.29          & 55.27          \\ \hline
\multirow{2}{*}{MORE}          & \multirow{2}{*}{VoteNet  Graph GA}        & 3D                    & 38.98          & 23.01          & 21.65          & 44.33          & 31.93 & 58.89          & 35.41          & 26.36          & 55.41          \\
                               &                                                    & 2D+3D                 & 40.94          & 22.93          & 21.66          & 44.42          & 33.75 & 62.91          & 36.25          & 26.75          & 56.33          \\ \hline
\multirow{2}{*}{X-Trans2Cap}   & \multirow{2}{*}{VoteNet  KD}                & 3D                    & 41.52          & 23.83          & 21.09          & 44.97          & 34.68 & 58.81          & 34.17          & 25.81          & 54.10          \\
                               &                                                    & 2D+3D                 & 43.87          & 25.05          & 22.46          & 45.28          & 35.31 & 61.83          & 35.65          & 26.61          & 54.70          \\ \hline
\multirow{2}{*}{SpaCap3D}      & \multirow{2}{*}{VoteNet  Transformer}              & 3D                    & 42.53          & 25.02          & 22.22          & 45.65          & 34.44 & 58.06          & 35.30          & 26.16          & 55.03          \\
                               &                                                    & 2D+3D                 & 44.02          & 25.26          & 22.33          & 45.36          & 36.64 & 63.30          & 36.46          & 26.71          & 55.71          \\ \hline
\multirow{2}{*}{3DJCG}         & \multirow{2}{*}{VoteNet  Transformer}                & 3D                    & 47.68          & 31.53          & 24.28          & 51.08          & -     & 60.86          & 39.67          & 27.45          & 59.02          \\
                               &                                                    & 2D+3D                 & 49.48          & 31.03          & 24.22          & 50.80          & -     & 64.70          & 40.17          & 27.66          & \textbf{59.23}          \\ \hline
\multirow{2}{*}{CM3D}          & \multirow{2}{*}{VoteNet  Transformer}              & 3D                    & 50.29          & 25.64          & 22.57          & 44.71          & 35.97 & -              & -              & -              & -              \\
                               &                                                    & 2D+3D                 & 54.30          & 27.24          & 23.30          & 45.81          & 42.77 & -              & -              & -              & -              \\ \hline
\multirow{2}{*}{D3Net}         & \multirow{2}{*}{PointGroup  Graph GA}                & 3D                    & -              & -              & -              & -              & -     & -              & -              & -              & -              \\
                               &                                                    & 2D+3D                 & 62.64          & 35.68          & \textbf{25.72}          & \textbf{53.90}          & \textbf{53.95} & -              & -              & -              & -              \\ \hline
\multirow{2}{*}{Vote2Cap-DETR} & \multirow{2}{*}{3DERT   Transformer}               & 3D                    & \textbf{73.77} & \textbf{38.21} & \textbf{26.64} & \textbf{54.71} & -     & \textbf{84.15} & \textbf{42.51} & \textbf{28.47} & \textbf{59.26} \\
                               &                                                    & 2D+3D                 & \textbf{70.63} & \textbf{35.69} & 25.51          & 52.28          & -     & \textbf{86.28} & \textbf{42.64} & \textbf{28.27} & 59.07              \\ \hline
\end{tabular}
\end{table*}

\begin{table}[h!]
\caption{Experimental results of different models on the Nr3D validation set. The best result reported in the original paper is taken as the comparison standard.}
\renewcommand\arraystretch{1.4}
\tabcolsep0.3cm
\centering
\label{tab4}
\begin{tabular}{c|c|cccc}
\hline
\multirow{2}{*}{Method} & \multicolumn{1}{l|}{\multirow{2}{*}{ Data}} & \multicolumn{4}{c}{IoU=0.5}                                       \\
                        & \multicolumn{1}{l|}{}                      & C              & B-4            & M              & R              \\ \hline
X-Trans2Cap             & 3D                                         & 30.96          & 18.70          & 22.15          & 49.92          \\
SpaCap3D                & 3D                                         & 31.43          & 18.98          & 22.24          & 49.79          \\
CM3D                    & 3D                                         & \textbf{35.86} & \textbf{20.73} & \textbf{22.86} & \textbf{51.23} \\ \hline
X-Trans2Cap             & 2D+3D                                      & 33.62          & 19.29          & 22.27          & 50.00          \\
SpaCap3D                & 2D+3D                                      & 33.71          & 19.92          & 22.61          & 50.50          \\
CM3D                    & 2D+3D                                      & 37.37          & 20.96          & 22.89          & 51.11          \\
3DJCG                   & 2D+3D                                      & 38.06          & 22.82          & 23.77          & 52.99          \\
D3Net                   & 2D+3D                                      & 38.42          & 22.22          & 24.74          & 54.37          \\
Vote2Cap-DETR           & 2D+3D                                      & \textbf{45.53} & \textbf{26.88} & \textbf{25.43} & \textbf{54.76} \\ \hline
\end{tabular}
\end{table}

\begin{table*}[]
\renewcommand\arraystretch{1.3}
\caption{The performance of the Scan2Cap online benchmark, the currently only benchmark that incorporates the test dataset of ScanRefer for 3D dense captioning. The superscripts in each column indicate the models' rankings based on different metrics. The first sorting criterion used is the CIDEr metric (C@0.5IoU). The overall ranking results are largely consistent with those in Table IV and Table V. However, a slight difference is observed where the individual D3Net-Speaker model is not as effective as the combined D3Net model.}
\centering
\renewcommand\arraystretch{1.5}
\tabcolsep0.3cm
\label{tab6}
\begin{tabular}{cccccccc}
\hline
\multirow{2}{*}{Method} & \multicolumn{5}{c}{Captioning}                            & Detection & Submission   \\ 
                        & C@0.5IoU  & B@0.5IoU  & R@0.5IoU  & M@0.5IoU  & DCmAP     & mAP@0.5   & date         \\ \hline
Vote2Cap-DETR\cite{chen2023end}           & \textbf{0.3128 $^{1}$ } & \textbf{0.1778 $^{1}$}  & \textbf{0.2842 $^{1}$}  & \textbf{0.1316 $^{1}$}  & \textbf{0.1825 $^{1}$}  & \textbf{0.4454 $^{1}$}  & 17 Nov, 2022 \\
CFM                     & 0.2360 $^{2}$  & 0.1417 $^{2}$  & 0.2253 $^{2}$  & 0.1034 $^{2}$  & 0.1379 $^{5}$  & 0.3008 $^{5}$  & 4 Nov, 2022  \\
CM3D\cite{zhong2022contextual}                    & 0.2348 $^{3}$  & 0.1383 $^{3}$  & 0.2250 $^{4}$  & 0.1030 $^{3}$  & 0.1398 $^{4}$  & 0.2966 $^{7}$  & 27 Sep, 2022 \\
Forest-xyz              & 0.2266 $^{4}$  & 0.1363 $^{4}$  & 0.2250 $^{3}$  & 0.1027 $^{4}$  & 0.1161 $^{10}$ & 0.2825 $^{10}$ & 6 Oct, 2022  \\
D3Net-Speaker\cite{takahashi2020d3net}           & 0.2088 $^{5}$  & 0.1335 $^{6}$  & 0.2237 $^{5}$  & 0.1022 $^{5}$  & 0.1481 $^{3}$  & 0.4198 $^{2}$  & 25 Aug, 2022 \\
3DJCG(Captioning)\cite{cai20223djcg}       & 0.1918 $^{6}$  & 0.1350 $^{5}$  & 0.2207 $^{6}$  & 0.1013 $^{6}$  & 0.1506 $^{2}$  & 0.3867 $^{3}$  & 12 Sep, 2022 \\
REMAN                   & 0.1662 $^{7}$  & 0.1070 $^{7}$  & 0.1790 $^{7}$  & 0.0815 $^{7}$  & 0.1235 $^{8}$  & 0.2927 $^{9}$  & 11 Sep, 2022 \\
NOAH                    & 0.1382 $^{8}$  & 0.0901 $^{8}$  & 0.1598 $^{8}$  & 0.0747 $^{8}$  & 0.1359 $^{6}$  & 0.2977 $^{6}$  & 28 Sep, 2022 \\
SpaCap3D\cite{wang2022spatiality}                & 0.1359 $^{9}$  & 0.0883 $^{9}$  & 0.1591 $^{9}$  & 0.0738 $^{9}$  & 0.1182 $^{9}$  & 0.3275 $^{4}$  & 31 Aug, 2022 \\
X-Trans2Cap\cite{yuan2022x}           & 0.1274 $^{10}$ & 0.0808 $^{11}$ & 0.1392 $^{11}$ & 0.0653 $^{11}$ & 0.1244 $^{7}$  & 0.2795 $^{11}$ & 29 Aug, 2022 \\
MORE-xyz\cite{jiao2022more}                & 0.1239 $^{11}$ & 0.0796 $^{12}$ & 0.1362 $^{12}$ & 0.0631 $^{12}$ & 0.1116 $^{12}$ & 0.2648 $^{12}$ & 11 Sep, 2022 \\
SUN+                    & 0.1148 $^{12}$ & 0.0846 $^{10}$ & 0.1564 $^{10}$ & 0.0711 $^{10}$ & 0.1143 $^{11}$ & 0.2958 $^{8}$  & 28 Sep, 2022 \\
Scan2Cap\cite{chen2020scanrefer}                & 0.0849 $^{13}$ & 0.0576 $^{13}$ & 0.1073 $^{13}$ & 0.0492 $^{13}$ & 0.0970 $^{13}$ & 0.2481 $^{13}$ & 25 Aug, 2022 \\ \hline
\end{tabular}
\end{table*}

\subsection{Performance Analysis}
The performance of the advanced methods on the ScanRefer and Nr3D datasets are summarized in Table \ref{tab3} and Table \ref{tab4}, respectively. Furthermore, Table \ref{tab6} presents the performance of the Scan2Cap online benchmark, the currently only benchmark that incorporates the test dataset of ScanRefer for 3D dense captioning. It is noteworthy to mention that 
we utilized the best outcomes reported in the original papers as our evaluation criterion, considering the presence of diverse data types and various training strategies presented for each approach.

Table \ref{tab3} summarizes the performance of various advanced methods on the ScanRefer dataset. The transformer-based framework Vote2Cap-DETR, equipped with innovative approaches, achieved state-of-the-art performance in terms of $C$ and $B-4$ index, regardless of whether the IoU threshold is set at 0.5 or 0.25. However, D3Net, another strong performer, outperformed Vote2Cap-DETR in terms of $M$ and $R$ index at IoU of 0.5 and achieved the best mAP result. It should be indicated that dual learning between 3D dense captioning and 3D visual grounding on larger datasets can significantly improve model performance.
In contrast, 3DJCG, a transformer-based model for dual learning tasks, performed worse than D3Net's non-transformer architecture. This may be due to the fact that 3DJCG used VoteNet as a scene encoder instead of replacing it with a more powerful detector, as done in D3Net, highlighting the importance of a strong scene encoder.
Transformer-based models such as CM3D, SpaCap3D, and X-Trans2Cap generally outperformed graph and GRU-based methods such as Scan2Cap and MORE, reflecting the dominance and potential of transformer-based approaches. Furthermore, incorporating a more comprehensive relationship module usually leads to performance improvement. For example, MORE, built on Scan2Cap, enhanced the relationship module and achieved a 3.78\% C@0.5IoU improvement over Scan2Cap in 3D input. Additionally, incorporating additional 2D input, such as pre-trained multi-view features, can further boost performance, which is evident in almost every method.

Table \ref{tab4} presents the performance of various methods on the Nr3D dataset, which is more challenging and focuses on fine-grained 3D object detection. The ranking trend of the methods' performance on Nr3D is similar to that on ScanRefer, with Vote2Cap-DETR remaining at the top of the list. However, the highest score achieved on Nr3D was only 45.53\% C@0.5IoU, which is quantitatively 25.1\% lower than the performance on ScanRefer. This indicates that existing models may struggle to provide distinctive descriptions when multiple instances of the same type of objects appear in a scene, which makes the task more challenging. Notably, 3DJCG showed superior performance compared to CM3D on the Nr3D dataset, suggesting that unified models that combine 3D dense captioning and 3D visual grounding could be more effective in dealing with complex scenes where fine-grained object detection is required. This highlights the importance of integrating different modalities and tasks to achieve better performance in challenging scenarios.

\section{Disscussion}\label{sec:disscussion}
Although prior works had made significant progress on 3D dense captioning, there remain several challenges that could be the focus of future studies. Hence, we will discuss the challenges and future trends of 3D dense captioning from six different aspects in this session, including datasets, external 2D knowledge, framework, generator module, vision-language pretraining technique, unified networks, and downstream applications.

\noindent\textbf{\itshape From Datasets}:
The limited availability of diverse and large-scale datasets for 3D dense captioning is a significant challenge. Existing datasets are based on indoor scenes and require human-annotated labels, which increases the cost of collecting data and introduces linguistic priors. Specifically, there are only two datasets \cite{chen2020scanrefer,achlioptas2020referit3d} for 3D dense captioning, while the number of image captioning datasets \cite{lin2014microsoft,young2014image,krishna2017visual,gan2017stylenet,gurari2020captioning,sharma2018conceptual,kuznetsova2020open,agrawal2019nocaps,mathews2016senticap} exceeds ten quantity.  Moreover, both datasets are based on indoor scenes, which inevitably carry more fixed relationships than random and complex outdoor scenes. In addition, most current models use supervised learning methods that require human-annotated labels for training, which greatly increases the cost of collecting the dataset. Moreover, it not only demands the accuracy of the dataset but also brings the problem of over-reliance on linguistic priors \cite{takahashi2020d3net}. Therefore,  future research can focus on developing unsupervised and reinforcement learning methods that rely less on labeled data and also explore ways to increase the diversity and size of datasets.

\noindent\textbf{\itshape From External 2D Knowledge}: The use of additional 2D features, such as those extracted by pre-trained ENet \cite{paszke2016enet}, is crucial for generating high-quality captions in 3D dense captioning. However, this can result in a computational burden \cite{yuan2022x}. Developing robust migration models that effectively integrate 2D and 3D features without compromising efficiency \cite{zhang2022learning} is a challenge that needs to be addressed. 

\noindent\textbf{\itshape From Framework}: Most of the previous work retains the detector, which causes the quality of the generated captions to be severely limited by the detection results. Although \cite{chen2023end} creates the precedent of the detector-free full end-to-end framework and obtains state-of-the-art performance, constructing a stronger detector-free model still needs exploration. Additionally, we can clearly observe from Table \ref{tab3} that the techniques we used are slightly homogeneous, which is either only graph-based or only transformer-based. Inspired by image captioning \cite{ghandi2022deep}, we can attempt to incorporate a diverse range of technologies in the future, such as combinations of graph-based and transformer-based techniques, and build more robust and effective models.

\noindent\textbf{\itshape From Generator Module}:  Most current generator decoders in 3D dense captioning generate sequential sentences word-by-word, which the previous word can influence. Exploring parallel word generation techniques, such as diffusion models, can enable bidirectional textual message interaction and potentially improve the generation process. It is worth mentioning that the diffusion model \cite{ho2020denoising} has made a significant breakthrough in visual content generation by generating the words in parallel and enabling bidirectional textual message interaction. Most recently, Chen \emph{et al.} confirmed the validity of the diffusion model on image captioning \cite{chen2022analog,luo2022semantic}, which may be able to be transferred to future 3D dense captioning tasks.

\noindent\textbf{\itshape From Vision-Language Pretraining Technique}: 
It is an undeniable fact that large-scale vision-language pre-training (VLP) models hold unparalleled advantages over other models, and their implementation in various fields has been steadily increasing. The recent release of GPT-4 \cite{openai2023gpt4} has particularly stirred the research community, creating a seismic impact. Notably, image captioning and dense image captioning \cite{mokady2021clipcap,hu2021vivo,li2020oscar,zhou2020unified} have witnessed a proliferation of relevant studies with remarkable results, leveraging models such as CLIP \cite{radford2021learning}, BERT \cite{devlin2018bert}, and GPT-2 \cite{radford2019language}. In light of these developments, it is imperative to bridge the gap in VLP for 3D dense captioning, which we believe will usher this research field into a new era.

\noindent\textbf{\itshape From Unified Networks and Downstream Applications}: 
Currently, an increasing number of researchers have expanded their scope beyond single-task exploration and placing greater emphasis on the integration of multiple tasks \cite{long2023capdet,gao2022caponimage,long2023capdet,gao2022caponimage}. Notably, the combination of 3D dense captioning with 3D visual grounding has been shown to possess significant potential \cite{takahashi2020d3net,cai20223djcg}. Consequently, the exploration of joint models for related tasks is emerging as a promising research direction, presenting both opportunities and challenges for the future and potentially giving rise to new 3D multimodal tasks.
Furthermore, there is a notable gap in the existing literature in terms of the limited attention given to downstream applications. The application of 3D dense captioning to assist individuals with visual impairments, for instance, holds greater relevance to real-world scenarios compared to 2D-based tasks. Furthermore, there are still unexplored opportunities and untapped potential in this area, awaiting further exploration and investigation.

\section{Conclusions}\label{sec:conclusion}
This paper presents a comprehensive review of the field of 3D dense captioning, which includes analyzing the main framework, datasets, related tasks, evaluation metrics, and future directions. By critically examining the strengths and weaknesses of existing models and considering the development trends in other domains, this paper has identified the current challenges and limitations of 3D dense captioning. The motivation of this paper is not only to provide a thorough understanding of the task for scholars who may be unfamiliar with it but also to serve as a source of inspiration and guidance for future research. It is our hope that this review will stimulate further investigations and advancements in the field of 3D dense captioning, ultimately leading to breakthroughs and advancements in this emerging research area.

\ifCLASSOPTIONcaptionsoff
  \newpage
\fi

\bibliographystyle{IEEEtran}
\bibliography{TCSVT2023}

\end{document}